\documentclass{article} %
\usepackage{iclr2026_conference,times}

\usepackage{amsmath,amsfonts,bm}

\def\eqref#1{equation~\ref{#1}}

\def\1{\bm{1}}

\DeclareMathAlphabet{\mathsfit}{\encodingdefault}{\sfdefault}{m}{sl}
\SetMathAlphabet{\mathsfit}{bold}{\encodingdefault}{\sfdefault}{bx}{n}

\usepackage{hyperref}
\usepackage{url}
\usepackage{graphicx}

\usepackage{amsmath}
\usepackage{amssymb}
\usepackage{multirow} 
\usepackage{wrapfig}
\usepackage{caption}
\usepackage[utf8]{inputenc} %
\usepackage{pifont}
\usepackage[T1]{fontenc}    %
\usepackage{hyperref}       %
\usepackage{url}            %
\usepackage{booktabs}       %
\usepackage{amsfonts}       %
\usepackage{nicefrac}       %
\usepackage{microtype}      %

\usepackage[table]{xcolor}         %
\usepackage{soul}

\usepackage[most]{tcolorbox}
\usepackage{lipsum}

\usepackage{rotating}

\definecolor{colorreldir}{HTML}{D8E2DC}
\definecolor{colorreldist}{HTML}{FFE5D9}
\definecolor{colorfindmy}{HTML}{FFCAD4}
\definecolor{colorafford}{HTML}{F4ACB7}
\definecolor{colorplan}{HTML}{ADA1A9}
\definecolor{colorsqa3d}{HTML}{6B9080}

\newcommand{\task}[2]{%
\begingroup%
\colorlet{hlcolor}{#1}%
\sethlcolor{hlcolor}%
\hl{#2}%
\endgroup%
}

\newcounter{MingyuanNumberOfComments}
\stepcounter{MingyuanNumberOfComments}

\newcounter{HaozhenNumberOfComments}
\stepcounter{HaozhenNumberOfComments}

\newcounter{BeitongNumberOfComments}
\stepcounter{BeitongNumberOfComments}

\newcommand{\dagdbl}{\texorpdfstring{\textsuperscript{\textdagger\textdaggerdbl}}{†‡}}

\title{Spatio-Temporal LLM: Reasoning about Environments and Actions}

\author{Haozhen Zheng, Beitong Tian, Mingyuan Wu, Zhenggang Tang, \\
\textbf{Klara Nahrstedt, Alex Schwing}\\
University of Illinois Urbana-Champaign \\
\texttt{\{haozhen3,aschwing\}@illinois.edu} \\
}

\iclrfinalcopy %
\begin{document}

\maketitle

\begin{abstract}
Despite significant recent progress of Multimodal Large Language Models (MLLMs), current MLLMs are challenged by ``spatio-temporal'' prompts, i.e., prompts that refer to 1) the entirety of an environment encoded in a point cloud that the MLLM should consider; and simultaneously also refer to 2) actions that happened in part of the environment and are encoded in a short ego-centric video clip. However, such a holistic spatio-temporal understanding is important for agents operating in the real world. To address this challenge, we first develop a framework to collect a large-scale dataset. Using the collected ``Reasoning about Environments and Actions'' (REA) dataset, we show that recent MLLMs indeed struggle to correctly answer ``spatio-temporal'' prompts. Building on this dataset, we study two spatio-temporal LLM (STLLM) baselines: 1) STLLM-3D, which directly fuses point cloud, video, and text representations as inputs to the LLM; and 2) STLLM-Aligner, which aligns spatial context with video and text before LLM decoding. Both baselines aim to enhance spatial understanding of environments and temporal grounding of egocentric observations. On REA, the STLLM baselines outperform existing models, demonstrating the effectiveness of our designs. Code and data are available at \url{https://zoezheng126.github.io/STLLM-website/}.

\end{abstract}

\section{Introduction}\label{sec:intro}

Despite significant advances, current Multimodal Large Language Models (MLLMs)  struggle to combine 3D spatial understanding with temporal reasoning in video data. While MLLMs have addressed multi-view understanding~\citep{yeh2025seeing} and other spatial reasoning tasks~\citep{cheng2024spatialrgpt, yang2024thinking}, and while some MLLMs exhibit strong spatial understanding, they are often trained solely on static image-text data. They hence lack the ability to model temporal dynamics, such as action progression, causal dependencies, or event ordering. This reveals a fundamental gap, %
motivating the need to develop a comprehensive spatio-temporal understanding.

To develop this, recent efforts by \citet{zhu2024llava} utilize 3D positional embeddings to enhance the 2D features with spatial context. Further, 
\citet{liu2024oryx}; \citet{liu2024coarsecorrespondencesboostspatialtemporal, li2024llavaonevisioneasyvisualtask} have enabled models to reason beyond 2D images and toward richer spatial representations by extending existing datasets to incorporate spatial data. Despite these advances, existing MLLMs see the world one frame at a time, i.e., grounded in the moment, and remain blind to the surrounding space. 
Notable exceptions \citep{man2024situational, huang2023embodied, ma2023sqa3dsituatedquestionanswering} extend Vision-Language Models for 3D situational awareness. These works incorporate additional 3D representations such as a point cloud or depth as an input. This enables agents to localize themselves in a scene %
and respond to spatial queries. %
However, these methods rely solely on static scene observations and do not incorporate temporal understanding, limiting their ability to reason about dynamic events or evolving interactions. %

In contrast, we envision a model that not only reasons about an unfolding event observed from an egocentric perspective, but also understands it from an allocentric view, anchoring temporal local motion in a broader world context. 
This is crucial for  embodied AI, situational awareness, and spatio-temporal question answering. E.g., robots interacting in the real world must interpret observations, not only by recognizing the current action, but by situating it within the surrounding context. %

\begin{figure}[t!]
    \centering
    \includegraphics[width=1.0\textwidth]{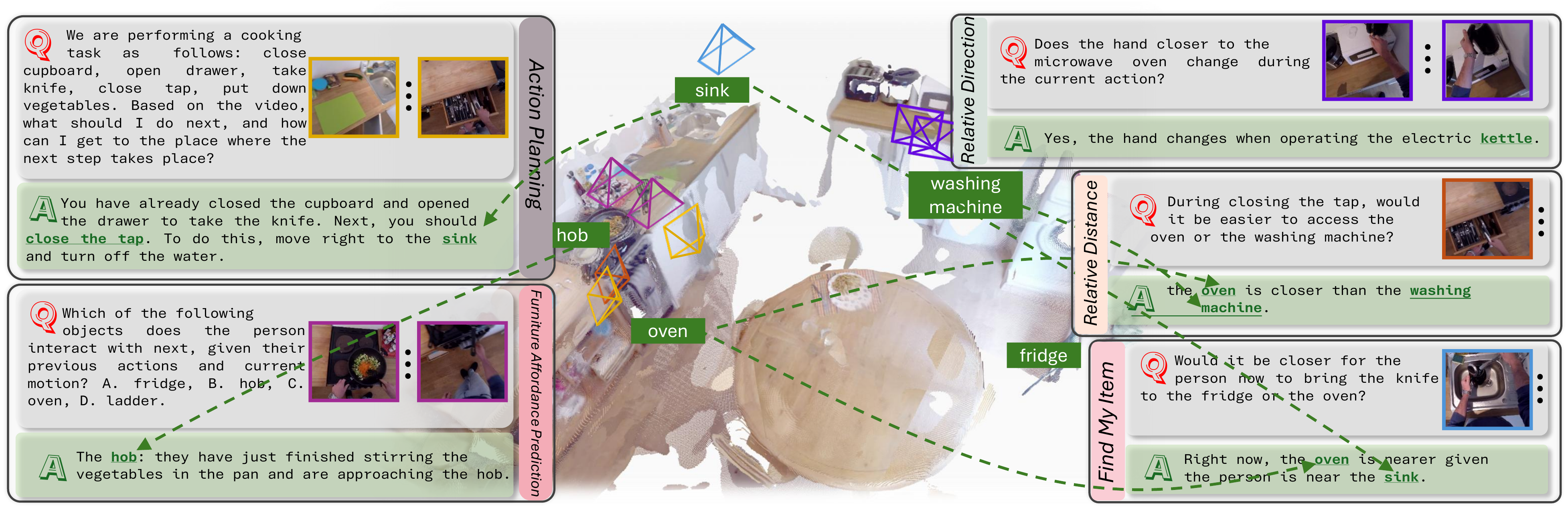}
    \vspace{-5mm}
    \caption{Spatial \emph{and} temporal reasoning is needed to answer prompts in ``Reasoning about Environments and Actions'' (REA). Ego-centric videos only show part of the point cloud environment.}
    \label{fig:teaser}
    \vspace{-3mm}
\end{figure}

To study joint spatial and temporal reasoning within a specified environment, we first develop a Spatio-Temporal Understanding Question Answering (QA) data collection pipeline. It is built upon work by \citet{damen2018scaling}, a widely recognized benchmark for temporal understanding. Using our pipeline, we collect the \textbf{``Reasoning about Environments and Actions'' (REA) data} shown in Fig.~\ref{fig:teaser}: it includes five tasks---relative direction, relative distance, find-my-item, furniture affordance prediction, action planning---each designed to test different aspects of spatio-temporal reasoning. 
We also study two \textbf{``Spatio-Temporal LLM'' (STLLM)} baselines %
which enhance spatio-temporal understanding by incorporating structured spatial knowledge into a video-language model. %

On REA data, we show: 1) spatio-temporal understanding remains a challenge for current MLLMs, as existing models achieve an overall \texttt{ChatGPT-4o}~\citep{gpt4o} LLM Judge accuracy of only 23.85\% to 31.46\% across tasks; and 2) our STLLM baselines reach 41.89\% overall accuracy and 47.32\% average categorical accuracy, highlighting that spatial and temporal cues are important.

In summary, our contributions are as follows: 1) we develop a dataset collection pipeline and collect REA (see Sec.~\ref{sec:data} for more); and 2) we study STLLM baselines to enhance spatio-temporal understanding in language models (see Sec.~\ref{sec:method} for more). %

\section{Related Work}\label{sec:related}

\noindent\textbf{Image LLM.} Recent commercial LLMs \citep{gpt4o,Llama3.2,Sonnet} have demonstrated strong results across a range of  image-language tasks, including image understanding and chart and task-based question answering. To mimic, most modern, open-source image-language models \citep{molmo, Qwen-VL, Qwen2VL, bai2025qwen25vltechnicalreport, shi2024eagle, li2025eagle2buildingposttraining, tong2024cambrian1fullyopenvisioncentric} adopt a common architecture: an image encoder, a connector module that pools and projects visual features into the LLM's embedding space, and a language decoder. Post-training techniques such as visual instruction tuning \citep{liu2023llava} are often applied to further enhance these models' ability to understand and follow natural language instructions. 

\noindent\textbf{Video LLM.} Early video-language models \citep{damonlpsg2023videollama, damonlpsg2024videollama2} adopt the image LLMs architecture: encode video frame features and connect them to an instruction-tuned language model via a projection layer. Recent efforts pursue %
long-form videos \citep{wu2025longvideobench, chen2024longvila, xu2025slowfastllava15familytokenefficientvideo}, improve streaming efficiency for real-time applications \citep{qian2024streaming, zhang2024flash, VideoLLM-online}, %
and introduce memory %
to enable effective long-term grounding and downstream question answering \citep{wang2024lifelongmemory, mangalam2024egoschema}.

\noindent\textbf{3D LLM.} Recent works \citep{hong20233dllminjecting3dworld, chen2024spatialvlmendowingvisionlanguagemodels} also inject 3D information directly into LLMs.  Due to the limited availability of %
3D datasets aligned with text, %
models are typically not trained from scratch. Instead, most approaches extract 2D visual features and project them back into 3D representations, which are then aligned with an LLM's embedding space for spatial reasoning.

\noindent\textbf{Image+3D LLM.} Extending beyond text-only reasoning with 3D inputs, recent works \citep{linghu2024multimodalsituatedreasoning3d, huang2023embodied}  interleave multimodal queries. %
When the images in the query are captured from egocentric viewpoints of an agent in a 3D environment, these queries offer a more grounded and accurate depiction of the agent’s surroundings and help the downstream task. %

\noindent\textbf{Video+3D LLM.} 
A natural extension of Image+3D LLMs is reasoning over both egocentric video and 3D environments. %
Towards this, recent models such as Video-3D LLM \citep{zheng2024video3dllmlearningpositionaware} and LLaVA-3D \citep{zhu2024llava} %
treat videos as sets of multi-view images, largely ignoring the temporal dynamics inherent in video data. As a result, they are naturally ill-suited for tasks that require fine-grained spatial-temporal reasoning, such as ours. Moreover, many of these approaches require extra information such as per-frame depth maps, which are not available in our dataset, making them incompatible with our evaluation setting. Thus, these methods aren't part of our baselines.

\noindent\textbf{Data for spatial understanding.}  
Several recent benchmarks assess %
spatial understanding in multimodal vision-language models \citep{yeh2025seeing, fu2024videommefirstevercomprehensiveevaluation, yang2024thinking, zhang2025flatlandspaceteachingvisionlanguage, cheng2024spatialrgpt}. While these datasets provide important testbeds for spatial reasoning, and some take videos as input to evaluate cross-frame spatial relationships, they are not grounded in human actions, limiting their relevance for tasks involving human-object interactions. In contrast, we introduce a setting that requires two streams of visual input: one capturing the holistic structure of the environment, and the other encoding local, dynamic changes within the scene, linking global spatial context with fine-grained temporal cues essential for real-world embodied QA.

\section{Reasoning about Environments and Actions (REA)}\label{sec:data}
\begin{wrapfigure}{r}{0.3\textwidth}
  \centering
  \includegraphics[width=\linewidth]{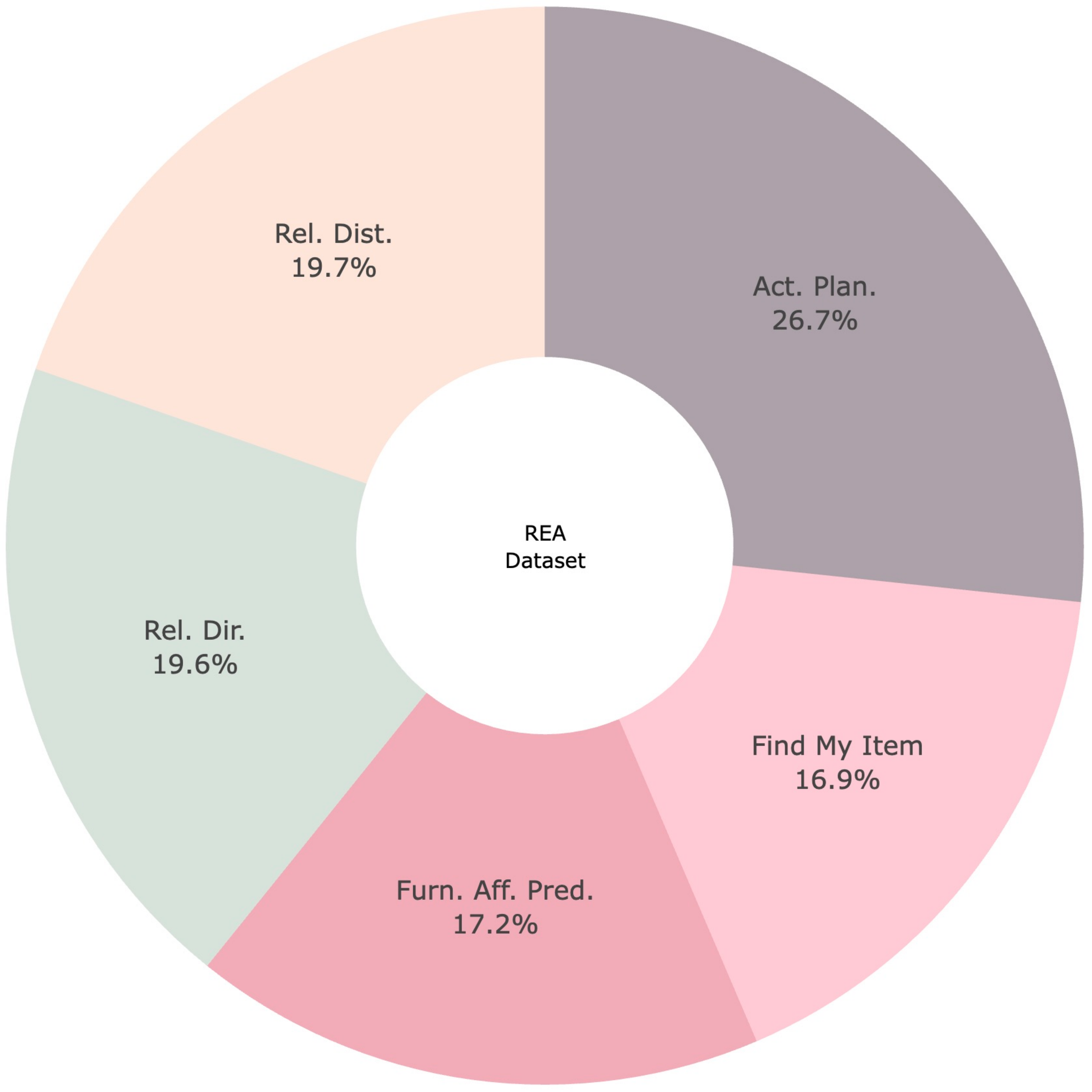}
  \captionsetup{font=small}
  \caption{Training data statistics.}
  \label{fig:data_stats}
\end{wrapfigure}

Our goal: equip MLLMs with spatial \emph{and} temporal understanding. %
As shown in Fig.~\ref{fig:teaser}, the model should answer prompts that require 1) spatial understanding about a \emph{global} 3D environment, represented via a point cloud; %
and 2) \emph{local} temporal understanding, represented via an egocentric video that covers part of the environment. %
For this we first develop a dataset collection pipeline benefitting from existing data: 
1) dense action annotations from EPIC-KITCHENS~\citep{damen2018scaling}; %
2) object segmentation annotations by VISOR~\citep{VISOR2022}; 
and 3) sparse point clouds from EPIC-FIELDS~\citep{EPICFields2023}. Using the pipeline, we collect the  \textbf{``Reasoning about Environments and Actions'' (REA)} data: question-answer pairs partitioned into five tasks that can be used to equip a MLLM with spatio-temporal understanding.  %
Answering prompts requires to understand
global scene context from a 3D scene  
and localized temporal cues from an egocentric video. Next, we first introduce the five tasks that form the REA dataset as shown in Fig.~\ref{fig:data_stats}, followed by details of the data collection pipeline (Sec.~\ref{subsec:pipeline}).

\setlength\intextsep{0pt}
\noindent\task{colorreldir}{\textbf{Relative Direction.}} 
This task requires to analyze  relative direction transitions of an object in the 3D scene during a series of actions performed by the person recording the egocentric video. The question asks to infer how the person's body orientation changes w.r.t.\ an object across two actions, such as inferring whether the hand closer to an object differs for two consecutive actions. The task can involve a single object or multiple objects. 
In the \textbf{single-object} setting, the question asks whether the person's movement or change in body orientation across a sequence of actions has resulted in a shift in the object's relative direction %
w.r.t.\ the person. 
In the \textbf{multi-object} setting, the question asks about the spatial relationship between multiple objects and the person, assessing whether their relative position or the person's viewpoint toward them remains consistent. %

\noindent\task{colorreldist}{\textbf{Relative Distance.}} This task evaluates the  ability to reason about how the person’s proximity to one or more objects changes over time, requiring spatial awareness across two query actions in the \textbf{single-object} setting and comparative distance understanding in the \textbf{multi-object} setting. Specifically, we ask questions such as ``\textit{Does the person move closer to the \textbf{query object} between the \textbf{query action 1} and \textbf{query action 2}?}'' (single-object), and ``\textit{During the first query action, is the person closer to the query object than to the reference object?}'' (multi-object).

\noindent\task{colorfindmy}{\textbf{Find My Item.}} The task assesses the  ability to localize an object %
and infer spatial %
steps %
to reach it, requiring integration of scene understanding and movement planning. An example question is: ``\textit{After performing the \textbf{query action}, where did the person leave the \textbf{query object} and how to reach it?}''. The model must identify the object's placement from the video and reason about the spatial path an agent would take to reach it.

\noindent\task{colorafford}{\textbf{Furniture Affordance Prediction.}} This task requires to predict which static furniture object the person is likely to interact with next, based on visual observations from the input video and spatial cues about the surrounding environment. The model must infer the recent action sequence, movement trend, and nearby layout. 
E.g., a question reads as ``\textit{Based on what the person has done so far and how they’re moving now, which nearby object is the person preparing to interact with?}''

\noindent\task{colorplan}{\textbf{Action Planning.}} This task evaluates the  ability to anticipate the next action in a task sequence and provide a navigation instruction to reach the location where the next step will occur. %

To collect data for all tasks, as illustrated in Fig.~\ref{fig:pcdreg}, 
we first sample a video clip containing past actions (Step 1) and compute the relative direction and distance to the anticipated next step (Step 2,3). To refine this spatial prediction, we include the video transition from the current action to the next (Step 4), allowing the model to reason about both motion and intent. Steps 5 and 6 incorporate 3D spatial cues by grounding video frames within the point cloud, enabling spatially-aware predictions of the next interaction location. We describe the six steps in Sec.~\ref{subsec:pipeline}.

\subsection{Data Collection Pipeline} \label{subsec:pipeline}

We obtain suitable question-answer (QA) pairs (Sec.~\ref{sec:qageneration}) and point cloud representations  (Sec.~\ref{sec:pointcloudgen}) in six steps as illustrated in Fig.~\ref{fig:pcdreg}. To ensure quality of the dataset, manual quality control was performed as discussed in Appendix~\ref{sec:A.1}.

\begin{figure}[t!]
    \centering
    \includegraphics[width=1.0\textwidth]{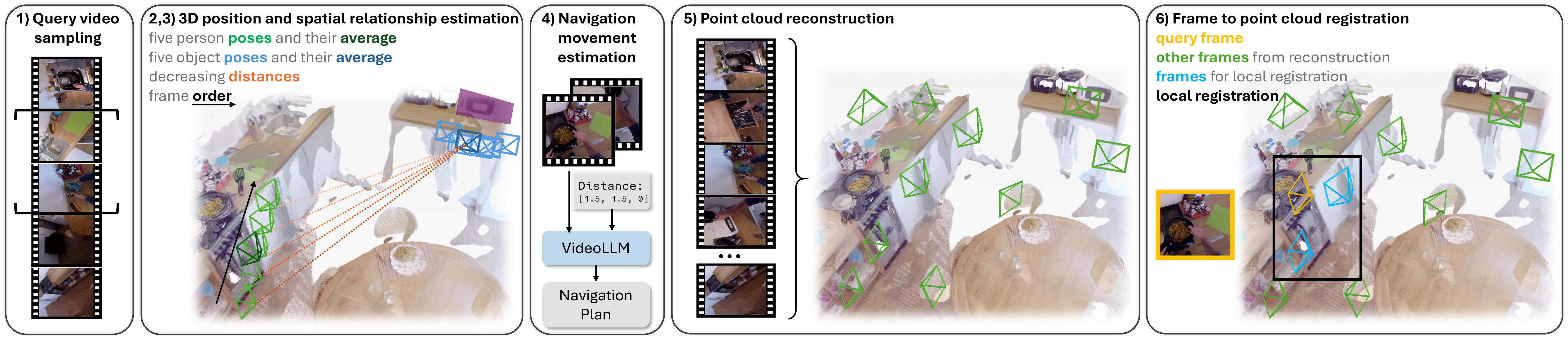}
    \vspace{-5mm}
    \caption{\textbf{Dataset generation pipeline.} Note, in 2\&3), camera poses (in green), sampled across the action interval, are used to compute the relative direction and distance between the person (moving along the arrow) and the object. 
    To obtain per-frame camera poses for the query video, we first use VGGT~\citep{wang2025vggtvisualgeometrygrounded} to re-compute the point cloud (step 5) and subsequently apply Reloc3r~\citep{dong2025reloc3rlargescaletrainingrelative} (step 6).
    }
    \label{fig:pcdreg}
    \vspace{-3mm}
\end{figure}

\subsubsection{Question-Answer (QA) Generation}  
\label{sec:qageneration}

To generate diverse question-answer pairs,  
we start with 3-5 question templates for each task which are rephrased via an LLM upon completion. 
To ensure that answering a question needs spatio-temporal understanding, completing a template requires the use of four steps: 1) query video sampling; 2) 3D position estimation; 3) spatial relationship estimation; and 4) navigation movement estimation.

\textbf{1) Query Video Sampling.} We begin by sampling 20-40 second clips, which serve as the visual input to the MLLM during training or inference. Clips are sampled from longer recordings in the EPIC-KITCHENS~\citep{damen2018scaling} data. Benefiting from the dense action annotations provided by EPIC-KITCHENS~\citep{damen2018scaling}, we partition a sampled clip into a series of fine-grained action intervals, each approximately 3-5 seconds long. We then sample action intervals from the clip. %
To be eligible, the selected actions must meet task-specific criteria. E.g., for
\task{colorreldir}{\emph{Relative Direction}} and \task{colorreldist}{\emph{Relative Distance}}, we select two non-consecutive actions from the clip with a sufficiently long interval between them, ensuring a high likelihood of a shift in relative angle or distance between the query object and the person.
In contrast, for \task{colorafford}{\emph{Furniture Affordance}} Prediction, where the goal is to infer which piece of furniture the person will interact with, the query actions are selected as the next action following the video clip. This ensures that the ground-truth interaction occurs after the clip ends, allowing the MLLM to predict the affordance without directly observing the final interaction.

\textbf{2) 3D Position Estimation.} Next, we compute the 3D locations of both the person and query objects. Given the short duration of each action interval and the typically small movements in kitchen scenarios, we assume that the person's location does not change significantly during a single action. To estimate the person's location during an action, we leverage the sparse image registration provided by EPIC-FIELD~\citep{EPICFields2023} and use the mean camera pose of the registered frames within the action interval as a proxy for the person's 3D location.

To determine the 3D location of a query object, we utilize the 2D segmentation masks from VISOR~\citep{VISOR2022}, along with the estimated human pose at the moment the person is interacting with the object. Assuming the object remains in close proximity to the person during the interaction, we project the 2D segmentation mask onto the COLMAP~\citep{schoenberger2016sfm} point cloud and compute the average 3D position of the projected points using a frame near the middle of the interaction. %
This average serves as the estimated 3D location of the query object.

\textbf{3) Spatial Relationship Estimation.} After estimating the ground-truth 3D poses of the person and the objects, we compute the spatial relationship between the object and the person's movement as observed in the query video. To simplify the analysis, we constrain the object to remain stationary throughout the movement by ensuring that the person interacts with the object at most once. %

For the \task{colorreldir}{\emph{Relative Direction}} task, computations are performed in the person's camera coordinate frame. By transforming the object's 3D location from world coordinates into the person's egocentric frame, we determine whether the object is positioned to the left, right, front, or back of the person.

For the \task{colorreldist}{\emph{Relative Distance}} task, we calculate the change in distance between the person and the object over time using their world-coordinate poses. Specifically, we sample five poses of the person across the action interval, compute the L2 distances to the object in world coordinates, and fit a linear regression to these values. A positive slope indicates the person is moving away from the object, while a negative slope indicates movement toward it. We apply a threshold of $\pm 0.05$ to classify whether the person is moving relative to the object or remains relatively stationary.

\textbf{4) Navigation Movement Estimation.} To estimate navigational movement and direction in the \task{colorplan}{\emph{Action Planning}} task, we begin by measuring displacement in world coordinates. Specifically, we apply a threshold of 1.5 meters to distinguish meaningful movement from minor body shifts or wobbling. We then compute the relative direction between the person’s current location and the destination to infer the intended navigation direction.

After obtaining this preliminary ground-truth estimate, we refine it using a VideoLLM~\citep{zhang2024videoinstructiontuningsynthetic}. The model is prompted with both the initial estimation and a reference video containing the ground-truth navigation sequence. The VideoLLM~\citep{zhang2024videoinstructiontuningsynthetic} is asked to observe the person's movement in the video and assess whether the preliminary result accurately reflects the true navigational behavior. This refinement step is crucial, as navigation involves not only directional displacement but also obstacle avoidance and complex scene understanding.

\vspace{-1mm}
\subsubsection{Point Cloud Reconstruction}
\label{sec:pointcloudgen}
\vspace{-1mm}
Although EPIC-FIELDS~\citep{EPICFields2023} provides 3D reconstructions for each scene in EPIC-KITCHENS~\citep{damen2018scaling}, it only offers sparse COLMAP~\citep{schoenberger2016sfm} models, containing limited camera poses sampled across full-length recordings. However, dense and accurate camera poses are crucial for complex question-answer tasks to capture both spatial layout and temporal context. To address this, we construct dense 3D models in two steps: first, we recompute point clouds using pose-free reconstruction methods~\citep{tang2024mv, wang2025vggtvisualgeometrygrounded}; then, we register dense camera poses from each query video using Reloc3r~\citep{dong2025reloc3rlargescaletrainingrelative}. This approach significantly reduces the time required for dense image-to-scene registration compared to COLMAP-based pipelines, while enabling high-fidelity scene grounding for downstream tasks.

\textbf{5) Point Cloud Reconstruction.} Careful selection of images is essential when generating point clouds from egocentric videos. To avoid introducing noise %
in the point cloud, we %
filter frames that contain hands using Grounded SAM2~\citep{jiang2024trex2, kirillov2023segany,ren2024grounded, ren2024grounding, liu2023grounding}. To enhance coverage of the reconstruction while maintaining computational efficiency, we apply K-Means clustering on the camera poses of the filtered frames and select 25 representative frames per recording for point cloud generation. Quality of the point clouds was manually verified.

\textbf{6) Frame to Point Cloud Registration.}
\label{sec:registration}
To obtain the camera pose for each video frame, we follow %
Reloc3r~\citep{dong2025reloc3rlargescaletrainingrelative}. 
For each scene, we construct a database containing the 25 frames used during point cloud reconstruction, along with their camera poses in the scene coordinate system. These poses are obtained from VGGT~\citep{wang2025vggtvisualgeometrygrounded} reconstruction. Given a new video frame, we retrieve the two spatially closest images from the database via image features and use VGGT~\citep{wang2025vggtvisualgeometrygrounded} to predict the relative camera poses between the video frame and the retrieved images. %
Hence, the video frame's camera pose is registered to the reconstructed point cloud.

\begin{figure}[t!]
    \centering
    \includegraphics[width=\textwidth, keepaspectratio]{fig/architectures-crop.pdf}
    \caption{Architectures of STLLM-3D and STLLM-Aligner.}
    \label{fig:architecture}
\end{figure}
\section{Spatio-Temporal LLMs (STLLMs)}\label{sec:method}

To handle spatio-temporal reasoning, we want spatio-temporal LLMs to process a global 3D point cloud $\mathbf{P}$, a video $\mathbf{V}$, a textual instruction $\mathbf{T}$, and camera parameters for each video frame, including intrinsics and extrinsics. 
Importantly, the point cloud $\mathbf{P}$ offers a \emph{global} 3D context, 
and the video input $\mathbf{V}$ records egocentric human actions situated locally within this environment.
Concretely, the point cloud is defined as $\mathbf{P} = \{ [\mathbf{p}_{xyz}, \mathbf{f}_{\text{rgb}}] \} \in \mathbb{R}^{N \times 6}$, where $\mathbf{p}_{xyz} \in \mathbb{R}^{3}$ are the 3D coordinates of each point and $\mathbf{f}_{\text{rgb}} \in \mathbb{R}^{3}$ are the corresponding RGB colors. 
The video is represented as a set of $T$ image frames $\mathbf{V} = \{ \mathbf{I}_1, \mathbf{I}_2, \dots, \mathbf{I}_T \}$, where each frame $\mathbf{I}_t \in \mathbb{R}^{H \times W \times 3}$ is associated with its own intrinsic and extrinsic matrices $\mathbf{K}_t\in\mathbb{R}^{4\times 4}$ and $\mathbf{E}_t\in\mathbb{R}^{4\times 4}$.

Since the study of integration of both spatial and temporal data into LLMs is in its infancy, we assess two complementary baselines: %
1) \textbf{STLLM-3D} directly concatenates 3D features with video and text inputs for decoding; 
2) \textbf{STLLM-Aligner} extracts spatial queries via a cross-modal alignment module, which results in a compact spatial representation that is used as LLM input. 
Both baselines aim to capture global scene context while following fine-grained temporal dynamics for spatio-temporal reasoning. Architectures of STLLM-3D and STLLM-Aligner are illustrated in Fig.~\ref{fig:architecture}. %
Both baselines are based on {LLaVA-Video-Qwen2}~\citep{zhang2024videoinstructiontuningsynthetic}, 
 but extend it to handle 3D spatial information in addition to video and text. Both share the same vision and point cloud encoder.  
 For the vision encoder, following {LLaVA-Video-Qwen2}~\citep{zhang2024videoinstructiontuningsynthetic}, we adopt SigLip \citep{zhai2023sigmoidlosslanguageimage} as our vision encoder. 
 For the point cloud encoder, given a dense point cloud ${\mathbf P}$, we first apply voxel-based downsampling to obtain a reduced set of representative points $\tilde{\mathbf{P}}$ before extracting point-wise feature embeddings $\mathbf{f}_{\text{pcd}}$ using a masked transformer decoder $\mathcal{T}_{\text{pcd}}$, 
i.e.,
$
\mathbf{f}_{\text{pcd}} = \mathcal{T}_{\text{pcd}}(\tilde{\mathbf{P}}) \in \mathbb{R}^{N \times 768}
$~\citep{peng2023openscene}. 
See Appendix~\ref{sec:D} for training details.

\noindent\textbf{STLLM-3D.} 
The STLLM-3D baseline directly integrates 3D information with video and text. 
After extracting point-wise features $\mathbf{f}_{\text{pcd}}$, we apply Farthest Point Sampling and grouping to form a set of compact spatial features. 
These features are projected into the language space via an MLP layer and concatenated with the image and text embeddings, 
which are then fed directly into the LLM decoder. 
Beneficially, STLLM-3D adopts a straightforward design that concatenates spatial, visual, and textual features for direct decoding. This simplicity makes the architecture easy to implement and computationally lightweight. However, given large and complex scenes, more token embeddings are required, which increases the LLM input and its computational cost. %
In other words, while STLLM-3D is effective at direct integration, it struggles with scalability. This limitation motivates the design of STLLM-Aligner, which seeks to compress spatial features into a compact  representation. %

\noindent\textbf{STLLM-Aligner.} 
The STLLM-Aligner baseline  introduces a cross-modal alignment module to bridge frozen pointcloud features, video, and text. 
A set of learnable queries are used to attend to  image and text features, and spatial context from the point cloud is incorporated through cross-attention. 
The resulting spatial queries, together with the image and text embeddings, are passed into the LLM decoder. 
This design provides a compact yet informative representation of the 3D scene. %
However, the token compression introduced by the alignment mechanism is harder to interpret. %
Hence, STLLM-Aligner trades the efficiency challenge of STLLM-3D with its own limitations. To understand the trade-offs, we study both baselines. %

For the STLLM-Aligner, we also study use of a high-frequency positional encoding in the alignment module as a complementary enhancement. The positional encoding provides geometric cues beyond raw coordinates. 
Concretely, for each video frame, we back-project pixels using camera intrinsics and extrinsics to obtain per-pixel ray directions, which are then normalized and downsampled to match the patch-level tokens from the vision encoder. 
For the point cloud, each 3D point is transformed into the first camera frame, and its normalized direction from the camera origin is taken as its ray vector.  
These ray directions are then mapped through a high-frequency encoding function and projected with a lightweight MLP to align dimensions with the modality features. 
The resulting position-aware features are fused with the original image and point cloud embeddings before entering the alignment module. 
This design complements the frozen encoders with explicit geometric cues. %

\section{Experiments}\label{sec:exp}

We now examine the effectiveness of our dataset and the baselines: 1) We compare to several state-of-the-art VideoLLMs, including the base model we pretrained from. For a fair comparison, we feed the global scene context to existing models in a multi-view image format, as they cannot directly process point cloud inputs. 2) We evaluate using standard question answering metrics and two LLM-Judges~\citep{llmjudge} to ensure consistency in reasoning and correctness.

\noindent\textbf{Dataset Statistics.} We construct the \textbf{REA} dataset using our proposed data generation pipeline. After manual validation, we obtain 24,371 training samples and 1,757 validation samples. The dataset inherits the action classes and over 300 annotated objects from EPIC-KITCHENS~\citep{damen2018scaling} and EPIC-FIELDS~\citep{EPICFields2023}, covering a wide range of kitchen activities. It features strong long-tail distributions in both training and test splits (e.g., 4,759 unique actions in training with over 20\% appearing only once), highlighting its richness and diversity.

\begin{table}[t]
\vspace{-2mm}
\centering
\caption{Comparison of models on various evaluation metrics. Sim = Sentence Similarity.}
\vspace{-2mm}
\small
\resizebox{\textwidth}{!}{
\begin{tabular}{l|c|c|c|c|c}
\toprule
\textbf{Model} & \textbf{Sim (\%)} $\uparrow$ & \textbf{CIDEr} $\uparrow$ & \textbf{BLEU (\%)} $\uparrow$ & \textbf{METEOR (\%)} $\uparrow$ & \textbf{ROUGE (\%)} $\uparrow$ \\
\midrule
LLaVA-Video-7B-Qwen2~\citep{zhang2024videoinstructiontuningsynthetic} & 65.83 & 20.79 & 10.25 & 19.68 & 23.71 \\
\midrule
LLaVA-OV-Qwen2-7B~\citep{li2024llavaonevisioneasyvisualtask} & 64.51 & 3.34 & 11.22 & 19.53 & 23.84 \\
\midrule
Qwen2-VL-7B-Instruct~\citep{Qwen2VL} & 52.99 & 35.08 & 19.11 & 17.38 & 25.37 \\
\midrule
VideoLLaMa3
~\citep{damonlpsg2025videollama3} & 39.14 & 10.85 & 2.15 & 7.85 & 14.12 \\
\midrule[1.5pt]
\multicolumn{6}{l}{\small\textit{Finetuned on REA dataset}} \\
\midrule
LLaVA-Video-7B-Qwen2$^{\dagger}$ & 85.26 & 387.72 & 60.34 & 43.18 & \textbf{72.09} \\
\midrule
LLaVA-OneVision-Qwen-7B$^{\dagger}$ & 85.09 & 400.23 & 61.90 & 42.41 & 71.11 \\
\midrule
STLLM-Aligner$^{\ddagger}$ (w Pos.Enc.) & 71.34 & 170.63 & 39.46 & 28.54 & 50.02 \\
\midrule
STLLM-Aligner (w Pos. Enc.) & 85.70 & 406.54 & 61.90 & \textbf{44.16} & \textbf{72.09} \\
\midrule
\textbf{STLLM-Aligner} & 85.58 & \textbf{406.68} & \textbf{62.01} & 43.94 & 72.03 \\
\midrule
\textbf{STLLM-3D} & \textbf{85.99} & 405.48 & 61.99 & 44.04 & 72.07 \\
\bottomrule
\end{tabular}
}
\footnotesize{\textit{Note.} $^\dagger$ indicates the existing model is finetuned on our REA dataset. $^{\ddagger}$ LLM layers are not finetuned.}
\label{tab:table1}
\end{table}

\noindent\textbf{Metrics.} We adopt SenSim~\citep{reimers-2019-sentence-bert}, CIDEr~\citep{vedantam2015ciderconsensusbasedimagedescription}, BLEU-4~\citep{papineni2002bleu}, METEOR~\citep{banerjee2005meteor}, and ROUGE-L~\citep{lin2004rouge} following standard question answering and captioning evaluation protocols. We also employ LLM Judges~\citep{llmjudge} to better capture reasoning correctness. To better assess results, we employ two LLM judges \texttt{ChatGPT-4o} (C)~\citep{gpt4o} and \texttt{Gemini 2.0 Flash} (G)~\citep{gemini2025flash} which both assign discrete correctness labels (``Correct''/``Wrong''). Task-oriented prompts are carefully designed to explicitly instruct the judge to assess the validity of the underlying reasoning (Appendix~\ref{sec:B}). Importantly, the LLM Judges are not solving the spatio-temporal reasoning tasks themselves, but simply verify whether a model's prediction semantically matches the ground truth answer, which is a substantially simpler objective. 
Reliability of the judges is discussed in Appendix~\ref{sec:A.2}.

\noindent\textbf{Model Implementation.} STLLM-3D and STLLM-Aligner baselines are built upon {LLaVA-Video-7B-Qwen2}~\citep{zhang2024videoinstructiontuningsynthetic}. 
They are finetuned on our {REA} dataset for one epoch using AdamW~\citep{loshchilov2017decoupled} with a cosine learning rate scheduler and a max learning rate of $1\mathrm{e}{-4}$ (training details in Appendix~\ref{sec:D}). Finetuning is conducted on four NVIDIA H200 GPUs.

\begin{table*}[t]
\centering
\caption{LLM-Judge accuracy (\%, higher is better, C = \texttt{ChatGPT-4o}, G = \texttt{Gemini 2.0 Flash}).}
\vspace{-2mm}
\small
\resizebox{\textwidth}{!}{%
\setlength{\tabcolsep}{6pt}
\begin{tabular}{l
|c
|>{\columncolor{colorreldir}}c
|>{\columncolor{colorreldist}}c
|>{\columncolor{colorfindmy}}c
|>{\columncolor{colorafford}}c
|>{\columncolor{colorplan}}c
|c}
\toprule
\textbf{Model} & \textbf{} 
    & \textbf{Rel.~Dir.} 
    & \textbf{Rel.~Dist.} 
    & \textbf{Find My Item} 
    & \textbf{Affordance} 
    & \textbf{Action Plan.} 
    & \textbf{Overall / Avg.} \\
\midrule
\multirow{2}{*}{LLaVA-Video-7B-Qwen2~\citep{zhang2024videoinstructiontuningsynthetic}} 
    & C & 36.67 & 43.00 & 28.06 & 53.05 & 13.17 & 30.96 / 34.79 \\
    & G & 46.00 & 42.67 & 38.49 & 56.27 & 27.33 & 39.50 / 42.15 \\
\midrule
\multirow{2}{*}{LLaVA-OV-Qwen2-7B~\citep{li2024llavaonevisioneasyvisualtask}} 
    & C & 15.33 & 36.00 & 25.54 & 50.18 & 9.00 & 23.85 / 27.21 \\
    & G & 36.67 & 40.00 & 40.65 & 51.61 & 23.50 & 35.74 / 38.49 \\
\midrule
\multirow{2}{*}{Qwen2-VL-7B-Instruct ~\citep{Qwen2VL}} 
    & C & 38.33 & 9.67 & 15.47 & 40.50 & 15.00 & 24.38 / 23.68 \\
    & G & 36.67 & 10.00 & 23.02 & 41.22 & \textbf{33.67} & 29.94 / 27.90 \\
\midrule
\multirow{2}{*}{VideoLLaMa3~\citep{damonlpsg2025videollama3}} 
    & C & \textbf{57.00} & 42.00 & 20.86 & 39.43 & 10.00 & 31.46 / 35.86 \\
    & G & \textbf{70.33} & 38.33 & 42.45 & 39.07 & 13.33 & 36.03 / 40.70 \\
\midrule[1.5pt]
\multicolumn{8}{l}{\small\textit{Finetuned on REA dataset}} \\
\midrule
\multirow{2}{*}{LLaVA-Video-7B-Qwen2$^{\dagger}$} 
    & C & 40.67 & 61.00 & 36.69 & 61.65 & 11.83 & 36.99 / 42.37 \\
    & G & 44.00 & 61.00 & 56.12 & 57.35 & 20.50 & 42.92 / 47.79 \\
\midrule
\multirow{2}{*}{LLaVA-OV-Qwen-7B$^{\dagger}$} 
    & C & 41.00 & 66.00 & 32.73 & 59.86 & 15.33 & 38.19 / 42.98 \\
    & G & 47.00 & 66.00 & 47.12 & 55.91 & 23.50 & 43.65 / 47.91 \\
\midrule
\multirow{2}{*}{STLLM-Aligner$^{\ddagger}$ (w Pos.Enc.)} 
    & C & 56.67 & 49.67 & 28.78 & 63.80 & 7.50 & 35.38 / 41.26 \\
    & G & 39.67 & 48.33 & 48.20 & 55.91 & 14.17 & 36.37 / 41.26 \\
\midrule
\multirow{2}{*}{STLLM-Aligner (w Pos.Enc.)} 
    & C & 49.00 & 69.00 & \textbf{38.13} & 59.50 & \textbf{17.00} & 41.43 / 46.53 \\
    & G & 50.00 & 69.00 & 55.40 & 53.41 & 24.33 & 45.87 / 50.43 \\
\midrule
\multirow{2}{*}{\textbf{STLLM-Aligner}} 
    & C & 50.67 & \textbf{70.67} & 36.69 & 62.72 & 15.83 & \textbf{41.89} / \textbf{47.32} \\
    & G & 51.33 & \textbf{70.67} & 55.04 & 55.56 & 23.83 & \textbf{46.50} / 51.29 \\
\midrule
\multirow{2}{*}{\textbf{STLLM-3D}} 
    & C & 48.00 & 68.00 & 35.61 & \textbf{65.69} & 14.83 & 40.94 / 46.43 \\
    & G & 51.00 & 68.00 & \textbf{56.47} & \textbf{58.06} & 23.17 & 46.39 / \textbf{51.34} \\
\bottomrule
\end{tabular}}
\label{tab:task_accuracy_comparison}
\vspace{-2mm}
\begin{flushleft}
\end{flushleft}
\vspace{-5mm}
\label{tab:table2}
\end{table*}

\noindent\textbf{Results on Standard QA Metrics.} Table~\ref{tab:table1} reports sentence-level and n-gram metrics. 
Despite straightforward handling of scene and video data, STLLM baselines improve upon existing models, including those finetuned on REA. 
This shows: spatio-temporal reasoning is yet unsolved. %

\noindent\textbf{Results on LLM-Judge.} Table~\ref{tab:table2} reports LLM-Judge accuracy across the five tasks of the REA test set. As shown, off-the-shelf MLLMs are challenged by spatio-temporal reasoning, with overall accuracy below \textbf{31.46\%/39.50\%} (C/G). Interestingly, VideoLLaMA3~\citep{damonlpsg2025videollama3} achieves superior performance on the \task{colorreldir}{\emph{Relative Direction}} task, likely due to exposure to spatially-grounded data during pretraining. But it does not show consistent advantages on the other tasks. While finetuning on REA brings noticeable gains for all models, our STLLM baselines achieve noticeably higher performance. In particular, STLLM-Aligner attains \textbf{41.89\%/46.50\%}, outperforming the directly finetuned LLaVA-Video-7B-Qwen$^\dagger$ (\textbf{36.99\%/42.37\%}). This shows that developing spatio-temporal LLM architectures is beneficial and can yield gains over REA-finetuned 2D counterparts. We also assess use of additional positional encoding in the alignment module, but find little performance difference, likely because REA emphasizes question answering rather than explicit 3D grounding. Hence, positional encodings offer limited benefits. We provide additional analyses in Appendix~\ref{sec:full_exp}.

\noindent\textbf{Task difficulty.} Tasks such as \task{colorfindmy}{\emph{Find My Item}} and \task{colorplan}{\emph{Action Planning}} are generally more challenging, as they require open-ended answers and involve both spatial and temporal reasoning. Meanwhile, both \task{colorreldir}{\emph{Relative Direction}} and \task{colorreldist}{\emph{Relative Distance}} demand strong spatial understanding. Our STLLM baselines demonstrate superior overall performance with well-balanced results across tasks, highlighting their robust spatial-temporal reasoning capabilities.

\noindent\textbf{Judge comparison.} We observe that predictions evaluated by \texttt{ChatGPT-4o} and \texttt{Gemini 2.0 Flash} follow consistent overall trends. While \texttt{Gemini 2.0 Flash} reports higher absolute scores in more open-ended tasks such as \task{colorfindmy}{\emph{Find My Item}} and \task{colorplan}{\emph{Action Planning}} (around $\sim$1.5$\times$ that of \texttt{ChatGPT-4o}), the relative ordering across models remains largely stable. This indicates that both judges agree on the comparative ranking of methods, supporting the robustness of our evaluation.

\begin{table*}[t]
\centering
\caption{SQA3D Test Set - Correct Rate (\%) per Question Type (GPT / Gemini).}
\vspace{-2mm}
\small
\resizebox{\textwidth}{!}{%
\setlength{\tabcolsep}{6pt}
\begin{tabular}{l|c|c|c|c|c|c|c}
\toprule
\textbf{Model} & \textbf{What} & \textbf{Is} & \textbf{How} & \textbf{Can} & \textbf{Which} & \textbf{Other} & \textbf{Average} \\
\midrule
LLaVA-Video-7B-Qwen2~\citep{zhang2024videoinstructiontuningsynthetic} 
& 41.68 / 45.34
& 53.66 / 52.61 
& 20.62 / 22.16 
& 53.29 / 49.70 
& 40.12 / 48.84
& 46.61 / 47.88 
& 43.04 / 45.04 \\
\midrule
LLaVA-OV-Qwen2-7B~\citep{li2024llavaonevisioneasyvisualtask}
& 39.49 / 44.79
& 59.93 / 59.93
& 25.26 / 31.44
& 49.70 / 53.89
& 36.05 / 41.86
& 52.12 / 57.20
& 43.98 / 41.86 \\
\midrule
Qwen2-VL-7B-Instruct~\citep{Qwen2VL}
& 33.82 / 34.37 
& 51.92 / 51.92
& 21.65 / 17.53 
& 52.69 / 49.70
& 38.37 / 41.28 
& 50.00 / 50.85 
& 40.42 / 40.24 \\
\midrule
VideoLLaMa3~\citep{damonlpsg2025videollama3}
& 38.76 / 39.85 
& 56.10 / 63.07 
& 39.69 / 38.66 
& 47.31 / 55.69 
& 33.14 / 34.88 
& 52.54 / 54.66 
& 44.29 / 47.16 \\
\midrule[1.5pt]
\multicolumn{6}{l}{\small\textit{Finetuned on REA dataset}} \\
\midrule
LLaVA-Video-7B-Qwen$^\dagger$ 
& 48.26 / 50.82 
& 65.51 / 65.51 
& 45.36 / 46.91 
& \textbf{60.48} / 61.08 
& 44.19 / 40.70 
& 49.58 / 51.27 
& 52.03 / 53.03 \\
\midrule
LLaVA-OV-Qwen2-7B$^\dagger$ 
& 38.76 / 47.53 
& 56.10 / 62.72 
& 39.69 / 39.18 
& 47.31 / 49.10 
& 33.14 / 40.12 
& 52.54 / 50.00 
& 44.29 / 48.97 \\
\midrule
STLLM-Aligner$^{\ddagger}$
& 46.25 / 44.79 
& 56.79 / 62.37 
& 47.42 / 50.52 
& 52.10 / \textbf{73.05} 
& \textbf{47.09} / 39.53 
& 47.03 / 50.00 
& 49.10 / 51.78 \\
\midrule
STLLM-Aligner (w Pos. Enc.)
& 49.17 / 50.55
& 63.93 / 65.51
& 51.31 / 50.00
& 56.02 / 58.08
& 46.20 / 45.35
& 55.17 / \textbf{58.90}
& 53.32 / 54.62 \\
\midrule
\textbf{STLLM-Aligner}
& 49.73 / 51.74
& 65.16 / 64.46
& 50.52 / 48.97
& 58.68 / 58.08
& 50.00 / \textbf{50.00}
& 55.51 / 55.51
& 54.40 / 54.71 \\
\midrule
\textbf{STLLM-3D}
& \textbf{50.27} / \textbf{51.92}
& \textbf{66.55} / \textbf{66.90}
& \textbf{53.09} / \textbf{52.06}
& \textbf{60.48} / 62.28
& 46.51 / 43.02
& \textbf{57.20} / 58.05
& \textbf{55.21} / \textbf{55.65} \\
\bottomrule
\end{tabular}}
\label{tab:sqa3d_table}
\begin{flushleft}
\vspace{-1mm}
\footnotesize{\textit{Note.} The models do not generate exact-match answers by design. We incorporate LLM judges for evaluation, where an answer is considered correct if it expresses the same meaning as the ground truth.}
\end{flushleft}
\end{table*}

\textbf{Cross-Dataset Evaluation.} Our REA dataset provides effective supervision that significantly boosts the generalization of MLLMs on diverse 3D-related tasks. In particular, we conduct zero-shot evaluation on SQA3D~\citep{ma2023sqa3dsituatedquestionanswering}, a situated QA benchmark designed to assess scene understanding for embodied agents. As shown in Table~\ref{tab:sqa3d_table}, our STLLM baselines consistently outperform strong existing models, highlighting the effectiveness of spatio-temporal LLM designs. We also observe that models finetuned on REA (marked with $\dagger$) achieve clear improvements compared to their vanilla counterparts, demonstrating the transferability of our dataset to downstream 3D reasoning tasks. For instance, LLaVA-Video-7B-Qwen~\citep{zhang2024videoinstructiontuningsynthetic} improves its average accuracy from \textbf{43–45\%} to \textbf{52–53\%}, and LLaVA-OV-Qwen2-7B~\citep{li2024llavaonevisioneasyvisualtask} improves from \textbf{42–44\%} to \textbf{44–49\%}.  These gains highlight that REA training substantially enhances cross-dataset generalization, even for strong existing video-language models.

\setlength\intextsep{0pt}
\begin{wrapfigure}{r}{0.6\textwidth}
  \centering
  \includegraphics[width=\linewidth]{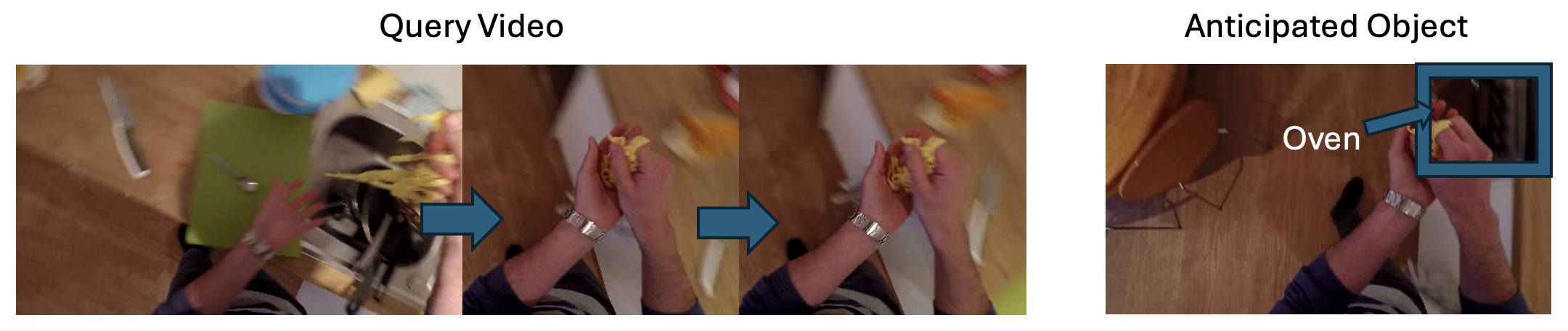}
  \captionsetup{font=small}
  \caption{Furniture Affordance Prediction example.}
  \label{fig:qualitative_result}
\end{wrapfigure}
\noindent\textbf{Qualitative results.} Fig.~\ref{fig:qualitative_result} 
presents a query video for a case from the \task{colorafford}{\emph{Furniture Affordance}} Prediction task, where the query asks: \textit{Which object will the person interact with next, the oven or the fridge?} The video shows the person completing an action at the sink, and the model must anticipate the next likely interaction. STLLM models correctly predict that \textit{“The person is preparing to interact with the oven, as they are moving closer to it,”} capturing the spatial intention toward a valid future object. In contrast, LLaVA-Video-7B-Qwen2~\citep{zhang2024videoinstructiontuningsynthetic} incorrectly responds that the person is \textit{“preparing to interact with the sink,”} a response grounded in the current frame rather than a forward-looking prediction. This highlights a key limitation of existing models: they often rely on immediate visual context without reasoning about temporal progression. In contrast, as intended spatio-temporal LLMs demonstrate compelling spatio-temporal reasoning.

\section{Conclusion}\label{sec:conc}

Joint spatio-temporal reasoning about an unfolding event given egocentric observations as well as a predicted allocentric environment representation, which permits to ground temporal local motion in a global context, is a crucial task for embodied AI. To study this capability, we collect the ``Reasoning about Environments and Actions'' (REA) data, consisting of five tasks, via a developed dataset collection pipeline. We show that classic multi-modal language models (MLLMs) struggle to correctly answer spatio-temporal prompts. We also show that results improve significantly when an MLLM is equipped with pointcloud understanding. %

\noindent\textbf{Limitations and broader impact.} While we present a first step towards joint spatio-temporal reasoning, much more work is needed to better understand 1) the type of training data that is most helpful for improving spatio-temporal reasoning of MLLMs; 2) the MLLM components that best extract meaningful information from the data. As a starting point for item 1 we present REA. As a starting point for item 2 we study two complementary baselines that highlight different trade-offs in extracting spatial information for multimodal reasoning. We envision significant positive broader impacts as spatio-temporal reasoning is a crucial component for embodied AI. However, successful spatio-temporal reasoning can also be abused for unnoticed mass surveillance. Hence, deployment of such technology requires great care. 

\noindent\textbf{Acknowledgements.} This work is supported in part by NSF grants 2008387, 2045586, 2106825, CNS 2106592, OAC 2126246, and NIFA award 2020-67021-32799. This research used the Delta advanced computing and data resource, which is supported by the National Science Foundation (award OAC 2005572) 
and the State of Illinois. The opinions expressed in this publication are those of the authors and do not necessarily reflect those of the funding agencies.

\bibliography{iclr2026_conference}
\bibliographystyle{iclr2026_conference}

\appendix
\newpage\clearpage
\section*{Appendix: Spatio-Temporal LLM: Reasoning about Environments and Actions}
This appendix is organized as follows: Sec.~\ref{sec:A} discusses the reliability of our dataset and the LLM-Judge metrics. Sec.~\ref{sec:B} provides the LLM Judge prompts used for evaluation. In Sec.~\ref{sec:C}, we include the inference instruction prompts used for baseline video models. Sec.~\ref{sec:D} outlines training details. Sec.~\ref{sec:full_exp} contains additional experiments and analyses. Sec.~\ref{sec:F} showcases additional qualitative results. Sec.~\ref{sec:G} provides further details about the \textbf{REA} dataset. Sec.~\ref{sec:limit} discusses  limitations. Finally, Sec.~\ref{sec:llm_usage} discusses the LLM usage in this work.

\section{Dataset and Metrics Reliability} \label{sec:A}

\subsection{Dataset} \label{sec:A.1}

\begin{table}[h]
\centering
\small
\caption{Human Evaluation on a Subset of the Test Split (values are accuracy in \%).}
\vspace{-1em}
\resizebox{\textwidth}{!}{%
\begin{tabular}{l|c|c|c|c|c|c|c}
\toprule
\textbf{Task} & \textbf{PCD Quality} & \textbf{QA Quality} & \textbf{Spatio-Temporal} & \textbf{Spatial Relation} & \textbf{Temporal Logic} & \textbf{Semantic Correctness} & \textbf{Clarity of Question} \\
\midrule
Relative Direction   & 80.00 & 80.00 & 80.00 & 85.00 & 100.00 & 95.00 & 100.00 \\
\midrule
Relative Distance    & 90.00 & 80.00 & 95.00 & 80.00 & 100.00 & 100.00 & 100.00 \\
\midrule
Find My Item         & 100.00 & 80.00 & 65.00 & 75.00 & 100.00 & 100.00 & 90.00 \\
\midrule
Furniture Affordance & 85.71 & 80.95 & 85.71 & 90.48 & 100.00 & 90.48 & 100.00 \\
\midrule
Action Planning      & 100.00 & 85.00 & 75.00 & 95.00 & 85.00 & 90.00 & 100.00 \\
\midrule
Overall              & 91.09 & 81.19 & 80.20 & 85.15 & 97.03 & 95.05 & 98.02 \\
\bottomrule
\end{tabular}}
\label{tab:human_study_table}
\end{table}

We implemented manual quality control during dataset collection to ensure the reliability of both the QA pairs and the VideoLLM-based refinement process. Specifically, we performed human verification and evaluation on the generated VQA pairs in the test set. We conducted the human evaluation on 100 samples (20 per task), where two expert annotators independently reviewed each sample. A score was marked as 1 only if both annotators agreed it was correct. We evaluate the test set using the following criteria:
\begin{itemize}
\item Point Cloud Quality: Does the point cloud accurately capture the positions of the reference objects?
\item QA Quality: Scored as 1 only if all of the following five sub-criteria are rated 1: spatial relation, temporal logic, semantic correctness, and clarity of the question; otherwise, 0.
\item Spatio-Temporal Reasoning: Does the question require understanding of both spatial layout and temporal action sequence? Does the answer demonstrate such reasoning?
\item Spatial Relation: Is the spatial relationship between the query object and the person accurately described in the answer?
\item Temporal Logic: Does the video contain the actions mentioned in the question in a temporally coherent manner?
\item Semantic Correctness: Does the answer correctly and clearly explain the reason behind the movement or action?
\item Clarity of Question: Is the question phrased clearly, fluently, and naturally, as if written by a human?
\end{itemize}

The REA dataset is specifically designed so that the query video often does not directly show the query objects, making point cloud information essential for spatial reasoning. As shown in Table~\ref{tab:human_study_table}, to answer correctly, 80\% of the questions require access to the point cloud.

\subsection{LLM-Judge} \label{sec:A.2}
We have incorporated human evaluation to refine the LLM Judge prompts. Specifically, after each prompt update, we randomly sampled 40 examples per task and compared the LLM Judge's verdicts with human judgments. We iteratively refined the prompt until the agreement between the LLM Judge and human evaluation exceeded 97\%. Please refer to Appendix~\ref{sec:B} for the final LLM Judge prompt. We note that the LLM Judge is particularly strict on tasks such as \task{colorfindmy}{\emph{Find My Item}} and \task{colorplan}{\emph{Action Planning}} as these involve open-ended question answering. This strictness helps ensure consistent and high-quality evaluation for tasks with less constrained answers. Also note that the LLM judge is asked to compare a given answer predicted by the models we study to the given ground truth answer. Hence, the LLM judge is not required to solve the spatio-temporal reasoning task. Instead, the LLM judge is tasked to assess the equivalence of the provided statements.

\section{LLM Judge Prompt} \label{sec:B}
We design task-specific prompts for the LLM Judge. The prompt is the same for \texttt{ChatGPT-4o} and \texttt{Gemini 2.0 Flash}.

\begin{tcolorbox}[
    enhanced,
    breakable,
    colback=yellow!10!white,
    colframe=colorreldir,
    title=Relative Direction,
    coltitle=black,
    fonttitle=\bfseries,
    listing only,
    listing options={
      basicstyle=\ttfamily\footnotesize,
      breaklines=true,
      showstringspaces=false
    }
]
You are a helpful and fair evaluator. Your task is to determine whether the predicted answer correctly follows the ground truth answer for a relative direction query. This task involves reasoning about the directional relationship between the person and a referenced object during two different actions, based on the scene.

\textbf{Predicted Result}: \{pred\}

\textbf{Ground Truth Result}: \{gt\}

\textbf{Query}: \{query\}

Please answer only with ``Correct'' or ``Wrong'', based on the following criteria:

- Mark as ``Correct'' if the predicted answer accurately reflects the relative direction of the object (e.g., left, right, forward, behind) in relation to the person across both mentioned actions, even if the wording differs.

- Directional terms may vary slightly (e.g., ``in front'' vs.\ ``forward'') but must preserve spatial meaning.

- The answer must address both actions in the query.

- If the prediction misidentifies one or both relative directions, or skips one, mark it as ``Wrong''.

Only reply with one word: ``correct'' or ``wrong'' — no explanation or extra text. If the prediction matches the ground truth, reply ``correct''. Otherwise, reply ``wrong''.
\end{tcolorbox}

\begin{tcolorbox}[
    enhanced,
    breakable,
    colback=yellow!10!white,
    colframe=colorreldist,
    title=Relative Distance,
    coltitle=black,
    fonttitle=\bfseries,
    listing only,
    listing options={
      basicstyle=\ttfamily\footnotesize,
      breaklines=true,
      showstringspaces=false
    }
]
You are a helpful and fair evaluator. Your task is to determine whether the predicted answer correctly follows the ground truth answer for a relative distance query. This task involves comparing the person’s distance to a specific object during two different actions, based on the scene.

\textbf{Predicted Result}: \{pred\}

\textbf{Ground Truth Result}: \{gt\}

\textbf{Query}: \{query\}

Please answer only with ``Correct'' or ``Wrong'', based on the following criteria:

- Mark as ``Correct'' if the predicted answer accurately conveys the relative distance relationship described in the ground truth, even if expressed with different wording.

- The prediction must clearly indicate which action places the person closer (or if the distances are about the same).

- Minor wording variations or additional clarifications are acceptable as long as the core spatial relationship is preserved.

- If the prediction contradicts or misses the comparison stated in the ground truth, mark it as ``Wrong''.

- Move closer and move further sometimes can be similar to remain the same distance, based on the context of the prediction, give reasonable judgement. 

Only reply with one word: ``correct'' or ``wrong'' — no explanation or extra text. If the prediction matches the ground truth, reply ``correct''. Otherwise, reply ``wrong''.
\end{tcolorbox}

\begin{tcolorbox}[
    enhanced,
    breakable,
    colback=yellow!10!white,
    colframe=colorfindmy,
    title=Find My Item,
    coltitle=black,
    fonttitle=\bfseries,
    listing only,
    listing options={
      basicstyle=\ttfamily\footnotesize,
      breaklines=true,
      showstringspaces=false
    }
]
You are a helpful and fair evaluator. Your task is to determine whether the predicted answer correctly follows the ground truth answer for a `Find My Item' query. This task requires identifying the location of a target object and describing how the person can get to it, based on the scene.

\textbf{Predicted Result}: \{pred\}

\textbf{Ground Truth Result}: \{gt\}

\textbf{Query}: \{query\}

Please answer only with ``Correct'' or ``Wrong'', based on the following criteria:

- Mark as ``Correct'' if the predicted answer matches the essential intent and meaning of the ground truth, even if phrased differently.

- The answer must correctly identify the item's location and provide a reasonable description of how to reach it.

- Minor differences in language, additional helpful navigation details, or alternative phrasing are acceptable if the overall meaning is consistent with the ground truth.

- If the predicted answer omits key information, misidentifies the item's location, or gives an implausible or unrelated navigation instruction, mark it as ``Wrong''.

Only reply with one word: ``correct'' or ``wrong'' — no explanation or extra text. If the prediction matches the ground truth, reply ``correct''. Otherwise, reply ``wrong''.
\end{tcolorbox}

\begin{tcolorbox}[
    enhanced,
    breakable,
    colback=yellow!10!white,
    colframe=colorafford,
    title=Furniture Affordance,
    coltitle=black,
    fonttitle=\bfseries,
    listing only,
    listing options={
      basicstyle=\ttfamily\footnotesize,
      breaklines=true,
      showstringspaces=false
    }
]
You are a helpful and fair evaluator. Your task is to determine whether the predicted answer correctly follows the ground truth answer for a furniture affordance query. This task involves reasoning about the person's past actions and current movement to infer which nearby object they are most likely preparing to interact with.

\textbf{Predicted Result}: \{pred\}

\textbf{Ground Truth Result}: \{gt\}

\textbf{Query}: \{query\}

Please answer only with ``Correct'' or ``Wrong'', based on the following criteria:

- Mark as ``Correct'' if the prediction correctly identifies the most likely object of interaction based on the query and provides a valid rationale aligned with the ground truth.

- The predicted object must match the correct option (e.g., ``oven'' or ``fridge'').

- Minor differences in phrasing or additional reasoning are acceptable as long as the predicted object is the same and the rationale is plausible.

- If the prediction identifies the wrong object or gives an unreasonable explanation, mark it as ``Wrong''.

Only reply with one word: ``correct'' or ``wrong'' — no explanation or extra text. If the prediction matches the ground truth, reply ``correct''. Otherwise, reply ``wrong''.
\end{tcolorbox}

\begin{tcolorbox}[
    enhanced,
    breakable,
    colback=yellow!10!white,
    colframe=colorplan,
    title=Action Planning,
    coltitle=black,
    fonttitle=\bfseries,
    listing only,
    listing options={
      basicstyle=\ttfamily\footnotesize,
      breaklines=true,
      showstringspaces=false
    }
]
You are a helpful and fair evaluator. Your task is to determine whether the predicted answer correctly follows the ground truth answer for an action planning query. The action planning task involves reasoning about sequences of actions in a cooking or assembly video, and determining what to do next and how to get there. 

\textbf{Predicted Result}: \{pred\}

\textbf{Ground Truth Result}: \{gt\}

\textbf{Query}: \{query\}

Please answer only with ``Correct'' or ``Wrong'', based on the following criteria:

- Mark as Correct if the predicted answer matches the intent and content of the ground truth, even if the wording is different. Reasonable paraphrasing is acceptable.

- The answer must identify the correct next step in the sequence, based on the context.

- It must also provide a plausible description of how to reach the location of the next step.

- Minor differences in phrasing or additional helpful details are acceptable, as long as the core actions are logically consistent with the ground truth.

- Avoid over-penalizing answers for surface-level differences if they preserve the meaning and ordering of actions.

Only reply with one word: ``correct'' or ``wrong'' — no explanation or extra text. If the prediction matches the ground truth, reply ``correct''. Otherwise, reply ``wrong''.
\end{tcolorbox}

\begin{tcolorbox}[
    enhanced,
    breakable,
    colback=yellow!10!white,
    colframe=colorsqa3d,
    title=SQA3D,
    coltitle=black,
    fonttitle=\bfseries,
    listing only,
    listing options={
      basicstyle=\ttfamily\footnotesize,
      breaklines=true,
      showstringspaces=false
    }
]
You are a helpful and fair evaluator. Your task is to determine whether the predicted answer correctly follows the ground truth answer for a furniture affordance query. This task involves reasoning about the person's past actions and current movement to infer which nearby object they are most likely preparing to interact with.

\textbf{Predicted Result}: \{pred\}

\textbf{Ground Truth Result}: \{gt\}

\textbf{Query}: \{query\}

Please answer only with "Correct" or "Wrong", based on the following criteria:

- Mark as "Correct" if the predicted answer expresses or implies the correct object or direction mentioned in the ground truth, even if phrased as a sentence, includes assistant prefixes, or contains extra context.

- Slight mismatches, rephrasings, or formatting issues (e.g., "The suitcase is under the bed." vs. "suitcase") are acceptable as long as the prediction clearly reflects the correct meaning.

- Mark as "Wrong" only if the answer refers to an entirely different object, contradicts the spatial context, or fails to address the question meaningfully.

Only reply with one word: "correct" or "wrong" — no explanation or extra text. If the prediction matches the ground truth, reply "correct". Otherwise, reply "wrong".

\end{tcolorbox}

\section{Inference Instruction Prompts for VideoLLMs} \label{sec:C}
As the existing VideoLLMs cannot take the point cloud as input, we use the 25 multi-view images (used for point cloud reconstruction) as a static scene description, and then input the same 32 query video frames as in our models.

To enable fair evaluation by an LLM judge, we prompt these models to generate full-sentence explanations rather than short answers like ``yes'' or ``no''. This ensures that the judge assesses answers based on reasoning rather than matching surface-level correctness.

Additionally, since these existing models were not trained to interpret the input as two separate streams (i.e., a static scene and a dynamic query video), we explicitly include this structure in the prompt to guide their attention accordingly.

\begin{tcolorbox}[
    enhanced,
    breakable,
    colback=yellow!10!white,
    colframe=red!75!black,
    title=Instruction Prompts,
    coltitle=black,
    fonttitle=\bfseries,
    listing only,
    listing options={
      basicstyle=\ttfamily\footnotesize,
      breaklines=true,
      showstringspaces=false
    }
]
The first 25 images provide multi-view observations of the current scene the person is in, while the next 32 frames depict egocentric actions—please refer to both to answer the question.

\textbf{<Image>} \{question\}

Give explanations and reasoning for your answer. Answer in detail, and be specific. Do not random guess. If you don't know, say `I don't know'.
\end{tcolorbox}

\section{Training Details} \label{sec:D}

Our training follows the standard next-token prediction objective, optimizing the token-wise cross-entropy loss over the LLM outputs. 
During training, modality-specific encoders remain frozen. 
We finetune the modality projectors and the LLM decoder (adapted with LoRA~\citep{hu2022lora}), with STLLM-Aligner additionally updating the alignment module (see Fig.~\ref{fig:architecture}).

All training was conducted using four NVIDIA H200 GPUs. The models were trained on the REA dataset for one epoch, which took approximately 6 hours. We adopt single-epoch training to balance efficiency and generalization. The LLM decoder is adapted using LoRA~\citep{hu2022lora} finetuning, enabling efficient parameter updates while mitigating overfitting to fixed answer templates and preserving the ability to generalize beyond rigid output structures.

For both models, the point cloud encoder is executed in \texttt{float32} precision, whereas the remaining components are trained in \texttt{bfloat16} for efficiency. \texttt{Nquery} stands for the number of learnable embeddings in the alignment module.

\begin{table}[h]
\centering
\vspace{1em}
\caption{Training hyperparameters for STLLM-Aligner and STLLM-Aligner (w Pos. Enc.) model.}
\vspace{-1em}
\small
\begin{tabular}{ll}
\toprule
\textbf{Parameter} & \textbf{Value} \\
\midrule
Gradient Accumulation Steps & 8 \\
Learning Rate               & $1 \times 10^{-4}$ \\
Weight Decay                & 0 \\
Precision$^*$                   & bfloat16 \\
Max Frames    & 32 \\
Voxel Size                  & 0.06 \\
Nquery                    & 1024 \\
\bottomrule
\end{tabular}
\label{tab:train1}
\end{table}

In Tab.~\ref{tab:train1}, \texttt{Nquery} denotes the number of tokens in the learnable query.
\begin{table}[h]
\centering
\vspace{1em}
\caption{Training hyperparameters for STLLM-3D.}
\vspace{-1em}
\small
\begin{tabular}{ll}
\toprule
\textbf{Parameter} & \textbf{Value} \\
\midrule
Gradient Accumulation Steps & 8 \\
Learning Rate               & $1 \times 10^{-4}$ \\
Weight Decay                & 0 \\
Precision$^*$                   & bfloat16 \\
Max Frames    & 32 \\
Voxel Size                  & 0.06 \\
Npoint                      & 1024 \\
Radius                      & 0.2 \\
Nsample                     & 64  \\
\bottomrule
\end{tabular}
\label{tab:train2}
\vspace{1em}
\end{table}

In Tab.~\ref{tab:train2}, \texttt{Npoint} denotes the number of center points sampled from the point cloud features. For each center point, a local neighborhood is defined by a specified \texttt{Radius}, and up to \texttt{Nsample} points are gathered within this radius to form a group.

\subsection{System Prompt}
We explicitly inform the model that the first \texttt{Nquery} visual tokens encode global spatio-temporal context, which it should pay special attention to during reasoning.
\begin{tcolorbox}[
    enhanced,
    breakable,
    colback=yellow!10!white,
    colframe=red!75!black,
    title=Instruction Prompts for REA,
    coltitle=black,
    fonttitle=\bfseries,
    listing only,
    listing options={
      basicstyle=\ttfamily\footnotesize,
      breaklines=true,
      showstringspaces=false
    }
]
The first 1024 tokens encode learnable queries representing objects and locations in the 3D scene. The following tokens represent egocentric video of recent actions. Use both to reason about spatial references and temporal context when answering.
\end{tcolorbox}

\section{Quantitative Results} \label{sec:full_exp}
\begin{table*}[ht]
\centering
\caption{LLM-Judge accuracy (\%, higher is better, C = \texttt{ChatGPT-4o}, G = \texttt{Gemini 2.0 Flash}).}
\vspace{-2mm}
\small
\resizebox{\textwidth}{!}{%
\setlength{\tabcolsep}{6pt}
\begin{tabular}{l
|c
|>{\columncolor{colorreldir}}c
|>{\columncolor{colorreldist}}c
|>{\columncolor{colorfindmy}}c
|>{\columncolor{colorafford}}c
|>{\columncolor{colorplan}}c
|c}
\toprule
\textbf{Model} & \textbf{} 
    & \textbf{Rel.~Dir.} 
    & \textbf{Rel.~Dist.} 
    & \textbf{Find My Item} 
    & \textbf{Affordance} 
    & \textbf{Action Plan.} 
    & \textbf{Overall / Avg.} \\
\midrule
\multirow{2}{*}{LLaVA-Video-7B-Qwen2~\citep{zhang2024videoinstructiontuningsynthetic}} 
    & C & 36.67 & 43.00 & 28.06 & 53.05 & 13.17 & 30.96 / 34.79 \\
    & G & 46.00 & 42.67 & 38.49 & 56.27 & 27.33 & 39.50 / 42.15 \\
\midrule
\multirow{2}{*}{LLaVA-OV-Qwen2-7B~\citep{li2024llavaonevisioneasyvisualtask}} 
    & C & 15.33 & 36.00 & 25.54 & 50.18 & 9.00 & 23.85 / 27.21 \\
    & G & 36.67 & 40.00 & 40.65 & 51.61 & 23.50 & 35.74 / 38.49 \\
\midrule
\multirow{2}{*}{Qwen2-VL-7B-Instruct ~\citep{Qwen2VL}} 
    & C & 38.33 & 9.67 & 15.47 & 40.50 & 15.00 & 24.38 / 23.68 \\
    & G & 36.67 & 10.00 & 23.02 & 41.22 & \textbf{33.67} & 29.94 / 27.90 \\
\midrule
\multirow{2}{*}{VideoLLaMa3~\citep{damonlpsg2025videollama3}} 
    & C & \textbf{57.00} & 42.00 & 20.86 & 39.43 & 10.00 & 31.46 / 35.86 \\
    & G & \textbf{70.33} & 38.33 & 42.45 & 39.07 & 13.33 & 36.03 / 40.70 \\
\midrule[1.5pt]
\multirow{2}{*}{LLaVA-Video-7B-Qwen2$^{\dagger}$} 
    & C & 40.67 & 61.00 & 36.69 & 61.65 & 11.83 & 36.99 / 42.37 \\
    & G & 44.00 & 61.00 & 56.12 & 57.35 & 20.50 & 42.92 / 47.79 \\
\midrule
\multirow{2}{*}{LLaVA-Video-7B-Qwen2\dagdbl (25 spatial tokens)} 
    & C & 29.33 & 53.67 & 29.50 & 61.29 & 8.33 & 31.42 / 36.42 \\
    & G & 31.86 & 53.67 & 46.04 & 56.63 & 19.33 & 37.48 / 41.52 \\
\midrule
\multirow{2}{*}{LLaVA-OV-Qwen-7B$^{\dagger}$} 
    & C & 41.00 & 66.00 & 32.73 & 59.86 & 15.33 & 38.19 / 42.98 \\
    & G & 47.00 & 66.00 & 47.12 & 55.91 & 23.50 & 43.65 / 47.91 \\
\midrule
\multirow{2}{*}{LLaVA-OV-Qwen2-7B\dagdbl (25 spatial tokens)}
    & C & 33.00 & 47.00 & 33.09 & 54.84 & 10.17 & 31.08 / 35.62 \\
    & G & 35.67 & 47.33 & 49.64 & 48.75 & 20.00 & 36.60 / 40.28 \\
\midrule
\multirow{2}{*}{STLLM-3D$^{\ddagger}$ (32 Nquery)} & C & 44.67 & 32.67 & 26.62 & 63.44 & 12.17 & 31.65 / 35.91 \\
    & G & 46.33 & 39.67 & 40.29 & 62.37 & 22.33 & 38.59 / 42.40 \\
\midrule
\multirow{2}{*}{STLLM-Aligner$^{\ddagger}$ (w Pos. Enc., 32 Nquery)} 
    & C & 40.33 & 46.00 & 34.89 & 60.22 & 13.83 & 34.55 / 39.05 \\
    & G & 45.00 & 47.67 & 48.20 & \textbf{60.22} & 19.00 & 39.50 / 44.02 \\
\midrule
\multirow{2}{*}{STLLM-Aligner$^{\ddagger}$ (w Pos. Enc.)} 
    & C & 56.67 & 49.67 & 28.78 & 63.80 & 7.50 & 35.38 / 41.26 \\
    & G & 39.67 & 48.33 & 48.20 & 55.91 & 14.17 & 36.37 / 41.26 \\
\midrule
\multirow{2}{*}{STLLM-Aligner (w Pos.Enc.)} 
    & C & 49.00 & 69.00 & \textbf{38.13} & 59.50 & \textbf{17.00} & 41.43 / 46.53 \\
    & G & 50.00 & 69.00 & 55.40 & 53.41 & 24.33 & 45.87 / 50.43 \\
\midrule
\multirow{2}{*}{\textbf{STLLM-Aligner}} 
    & C & 50.67 & \textbf{70.67} & 36.69 & 62.72 & 15.83 & \textbf{41.89} / \textbf{47.32} \\
    & G & 51.33 & \textbf{70.67} & 55.04 & 55.56 & 23.83 & \textbf{46.50} / 51.29 \\
\midrule
\multirow{2}{*}{\textbf{STLLM-3D}} 
    & C & 48.00 & 68.00 & 35.61 & \textbf{65.69} & 14.83 & 40.94 / 46.43 \\
    & G & 51.00 & 68.00 & \textbf{56.47} & 58.06 & 23.17 & 46.39 / \textbf{51.34} \\
\bottomrule
\end{tabular}}
\footnotesize{\textit{Note.} $^\dagger$ indicates the existing model is finetuned on our REA dataset. $^{\ddagger}$ LLM layers are not finetuned.}
\label{tab:full_rea_llm_judge_table}
\vspace{-1em}
\end{table*}

\vspace{1em}
\begin{table}[h]
\centering
\caption{Comparison of models on various evaluation metrics. Sim = Sentence Similarity.}
\vspace{-1em}
\small
\resizebox{\textwidth}{!}{
\begin{tabular}{l|c|c|c|c|c}
\toprule
\textbf{Model} & \textbf{Sim (\%)} $\uparrow$ & \textbf{CIDEr} $\uparrow$ & \textbf{BLEU (\%)} $\uparrow$ & \textbf{METEOR (\%)} $\uparrow$ & \textbf{ROUGE (\%)} $\uparrow$ \\
\midrule
LLaVA-Video-7B-Qwen2~\citep{zhang2024videoinstructiontuningsynthetic} & 65.83 & 20.79 & 10.25 & 19.68 & 23.71 \\
\midrule
LLaVA-OV-Qwen2-7B~\citep{li2024llavaonevisioneasyvisualtask} & 64.51 & 3.34 & 11.22 & 19.53 & 23.84 \\
\midrule
Qwen2-VL-7B-Instruct~\citep{Qwen2VL} & 52.99 & 35.08 & 19.11 & 17.38 & 25.37 \\
\midrule
VideoLLaMa3
~\citep{damonlpsg2025videollama3} & 39.14 & 10.85 & 2.15 & 7.85 & 14.12 \\
\midrule
LLaVA-Video-7B-Qwen2\dagdbl (25 spatial tokens) & 81.76 & 304.05 & 48.63 & 34.45 & 59.55 \\
\midrule
LLaVA-Video-7B-Qwen2$^{\dagger}$ & 85.26 & 387.72 & 60.34 & 43.18 & \textbf{72.09} \\
\midrule
LLaVA-OneVision-Qwen-7B\dagdbl (25 spatial tokens) & 81.06 & 297.79 & 45.43 & 33.94 & 58.45 \\
\midrule
LLaVA-OneVision-Qwen-7B$^{\dagger}$ & 85.09 & 400.23 & 61.90 & 42.41 & 71.11 \\
\midrule
STLLM-3D$^{\ddagger}$ (32 Nquery)  & 73.27 & 141.29 & 37.65 & 28.87 & 48.78 \\
\midrule
STLLM-Aligner$^{\ddagger}$ (w Pos.Enc.) & 71.34 & 170.63 & 39.46 & 28.54 & 50.02 \\
\midrule
STLLM-Aligner (w Pos. Enc.) & 85.70 & 406.54 & 61.90 & \textbf{44.16} & \textbf{72.09} \\
\midrule
\textbf{STLLM-Aligner} & 85.58 & \textbf{406.68} & \textbf{62.01} & 43.94 & 72.03 \\
\midrule
\textbf{STLLM-3D} & \textbf{85.99} & 405.48 & 61.99 & 44.04 & 72.07 \\
\bottomrule
\end{tabular}
}

\label{tab:full_caption_metrics}
\end{table}

\textbf{Finetuned Existing Models.} We pool the multi-view images in LLaVA-Video-7B-Qwen2$^\dagger$ and LLaVA-OV-Qwen-7B$^\dagger$ into token sequences similar in length to our model’s spatial queries (36 tokens per image) for a fair comparison. 

\textbf{Additional Experiments.} We conduct additional experiments to explore two factors: 1) the number of learnable spatial queries (\texttt{Nquery}), and 2) the number of trainable parameters.
For a fair comparison with a smaller number of spatial queries (32 \texttt{Nquery}), we also finetune strong baseline models, namely LLaVA-Video-7B-Qwen\dagdbl (25 spatial tokens) and LLaVA-OV-Qwen-7B\dagdbl (25 spatial tokens), both of which represent spatial information using one token per multi-view image (25 multi-view images in total).
As shown in Table~\ref{tab:full_rea_llm_judge_table}, STLLM-Aligner$^\ddagger$ (w/ Pos. Enc.), which uses 1024 \texttt{Nquery}, outperforms STLLM-Aligner$^\ddagger$ (w/ Pos. Enc., 32 \texttt{Nquery}), demonstrating that a larger number of spatial queries leads to better performance. However, the improvement is relatively modest. This motivates us to increase the number of trainable parameters. By further adapting the LLM layers with LoRA~\citep{hu2022lora}, performance improves substantially: overall accuracy rises from 35.38\% / 36.37\% (C/G) for STLLM-Aligner$^\ddagger$ (w/ Pos. Enc.) to 41.43\% / 45.87\% for STLLM-Aligner (w/ Pos. Enc.).
Under all settings, as shown in both Table~\ref{tab:full_rea_llm_judge_table} and Table~\ref{tab:full_caption_metrics}, our STLLM baselines outperform the direct finetuned counterparts, which showcase the effectiveness of our model designs.

\section{Qualitative Results} \label{sec:F}
In the qualitative results below, \textcolor{green}{\ding{51}} and \textcolor{red}{\ding{55}} indicate whether the prediction was marked as correct or incorrect by \texttt{ChatGPT-4o}, which serves as our LLM judge. Each figure shows two representative frames sampled from the query video, rendered below the reconstructed point cloud of the scene to provide spatial context for the queried actions.

\begin{tcolorbox}[
    enhanced,
    breakable,
    colback=yellow!10!white,
    colframe=colorreldir!75,
    title=Relative Direction,
    coltitle=black,
    fonttitle=\bfseries,
    listing only,
    listing options={
      basicstyle=\ttfamily\footnotesize,
      breaklines=true,
      showstringspaces=false
    }
]

\begin{wrapfigure}{r}{0.38\linewidth}
\vspace{-5pt}
\includegraphics[width=\linewidth, keepaspectratio]{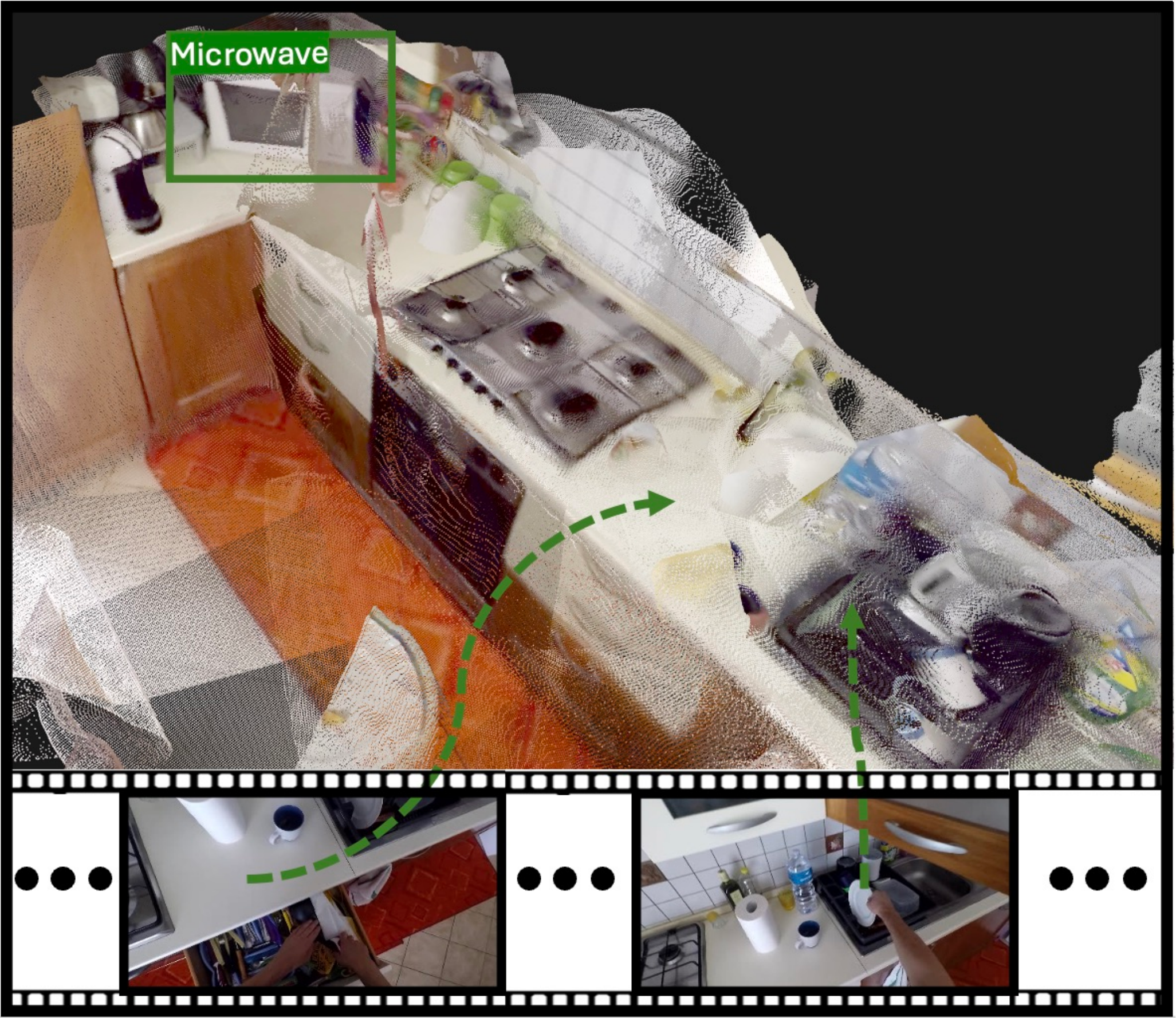}
\vspace{-5pt}
\end{wrapfigure}

\textbf{Question}: Is the microwave to the left of the person when the person is performing `put down knives' and also to the left of the person when performing `pick up plate'?  

\textbf{Ground-truth}: The microwave remains to the left of the person during both `put down knives' and `pick up plate'.

\textbf{Ours}: The microwave is to the left of the person before and after both `put down knives` and `pick up plate`. \textcolor{green}{\ding{51}} 

\textbf{LLaVA-OV-Qwen2-7B}: The microwave is not to the left of the person when they perform ``put down knives'' and ``pick up plate.'' The sequence shows that the person stands in front of the counter where the microwave is located, and they are not interacting with it during those specific actions. \textcolor{red}{\ding{55}}

\textbf{LLaVA-Video-7B-Qwen2}: The microwave is not to the left of the person when they are performing the action of putting down knives, as there are no knives visible in the images provided. The microwave is also not to the left of the person when they pick up a plate, as the person is seen reaching towards the sink area where the plate is located. \textcolor{red}{\ding{55}}

\textbf{Qwen2-VL-7B-Instruct}: No, the microwave is not to the left of the person when the person is performing `put down knives' and to the left of the person when `pick up plate'. \textcolor{red}{\ding{55}}
\end{tcolorbox}

\begin{tcolorbox}[
    enhanced,
    breakable,
    colback=yellow!10!white,
    colframe=colorreldist,
    title=Relative Distance,
    coltitle=black,
    fonttitle=\bfseries,
    listing only,
    listing options={
      basicstyle=\ttfamily\footnotesize,
      breaklines=true,
      showstringspaces=false
    }
]

\begin{wrapfigure}{r}{0.38\linewidth}
\vspace{-5pt}
\includegraphics[width=\linewidth, keepaspectratio]{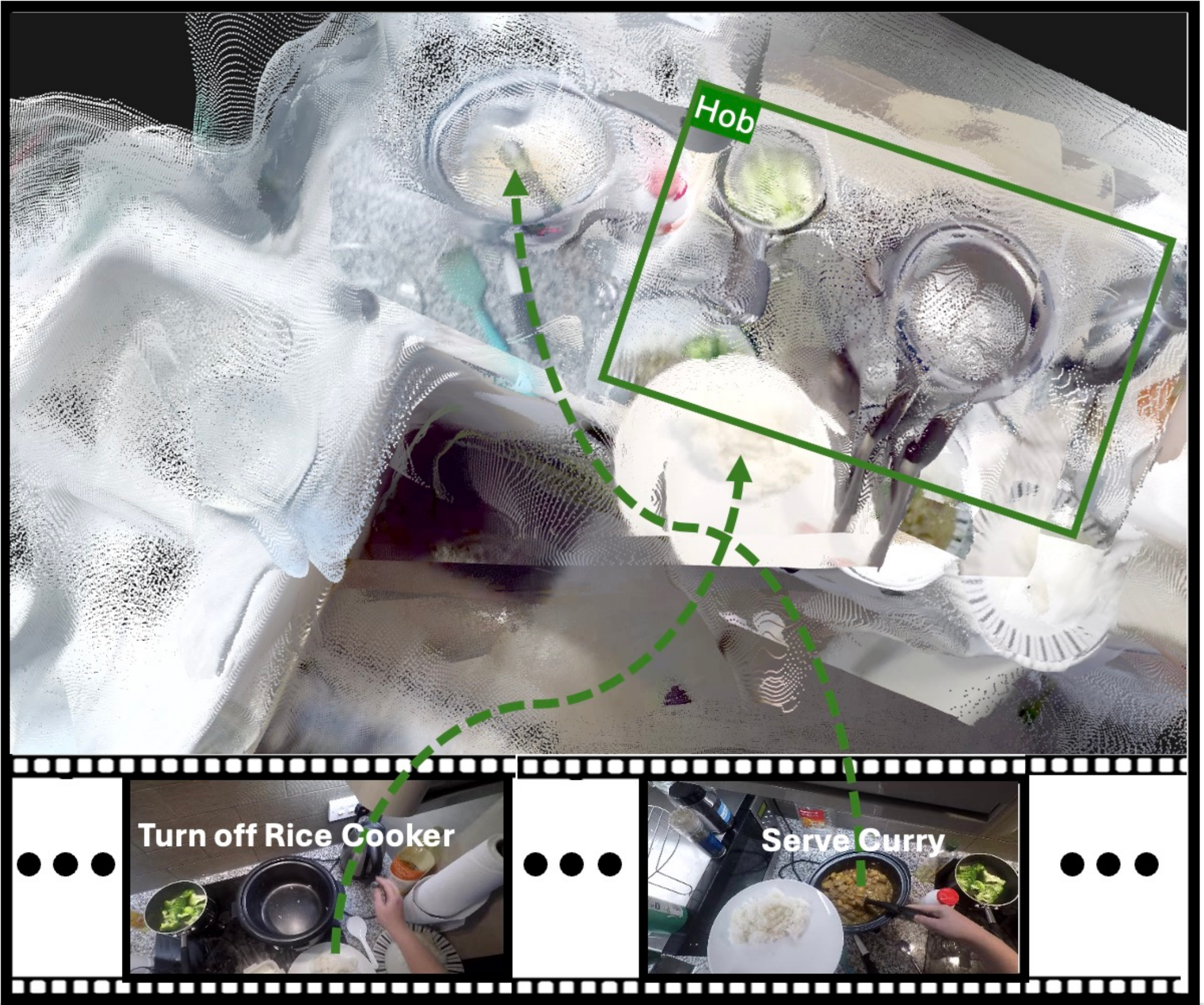}
\vspace{-5pt}
\end{wrapfigure}

\textbf{Question}: Does the person move closer to the hob between `turn off rice cooker' and `serve curry'?

\textbf{Ground-truth}: The person remains at about the same distance from the hob when performing both `turn off rice cooker' and `serve curry'.

\textbf{Ours}:  The person remains at about the same distance from the hob when performing both `turn off rice cooker' and `serve curry'. \textcolor{green}{\ding{51}} 

\textbf{LLaVA-OV-Qwen2-7B}: The person does not move closer to the hob between ``turn off rice cooker'' and ``serve curry''. The sequence of actions shows that after turning off the rice cooker, the person moves towards the sink area. The focus then shifts back to the countertop where the rice and curry are being served, but there is no direct interaction with the hob during these transitions. \textcolor{red}{\ding{55}} 

\textbf{LLaVA-Video-7B-Qwen2}: Yes, the person moves closer to the hob between `turn off rice cooker' and `serve curry'. The sequence of actions suggests that after turning off the rice cooker, the person is likely to be preparing to serve the dish. Moving closer to the hob implies that they are in the process of transferring the cooked curry onto the rice plates. \textcolor{red}{\ding{55}} 

\textbf{Qwen2-VL-7B-Instruct}: Yes, the person moves closer to the hob between `turn off rice cooker' and `serve curry'. In the first image, the person is standing further away from the hob, while in the second image, the person is standing closer to the hob. This can be inferred from the relative positions of the objects in the images. \textcolor{red}{\ding{55}} 
\end{tcolorbox}

\newpage
\begin{tcolorbox}[
    enhanced,
    breakable,
    colback=yellow!10!white,
    colframe=colorfindmy,
    title=Find My Item,
    coltitle=black,
    fonttitle=\bfseries,
    listing only,
    listing options={
      basicstyle=\ttfamily\footnotesize,
      breaklines=true,
      showstringspaces=false
    }
]
\begin{wrapfigure}{r}{0.38\linewidth}
\vspace{-5pt}
\includegraphics[width=\linewidth, keepaspectratio]{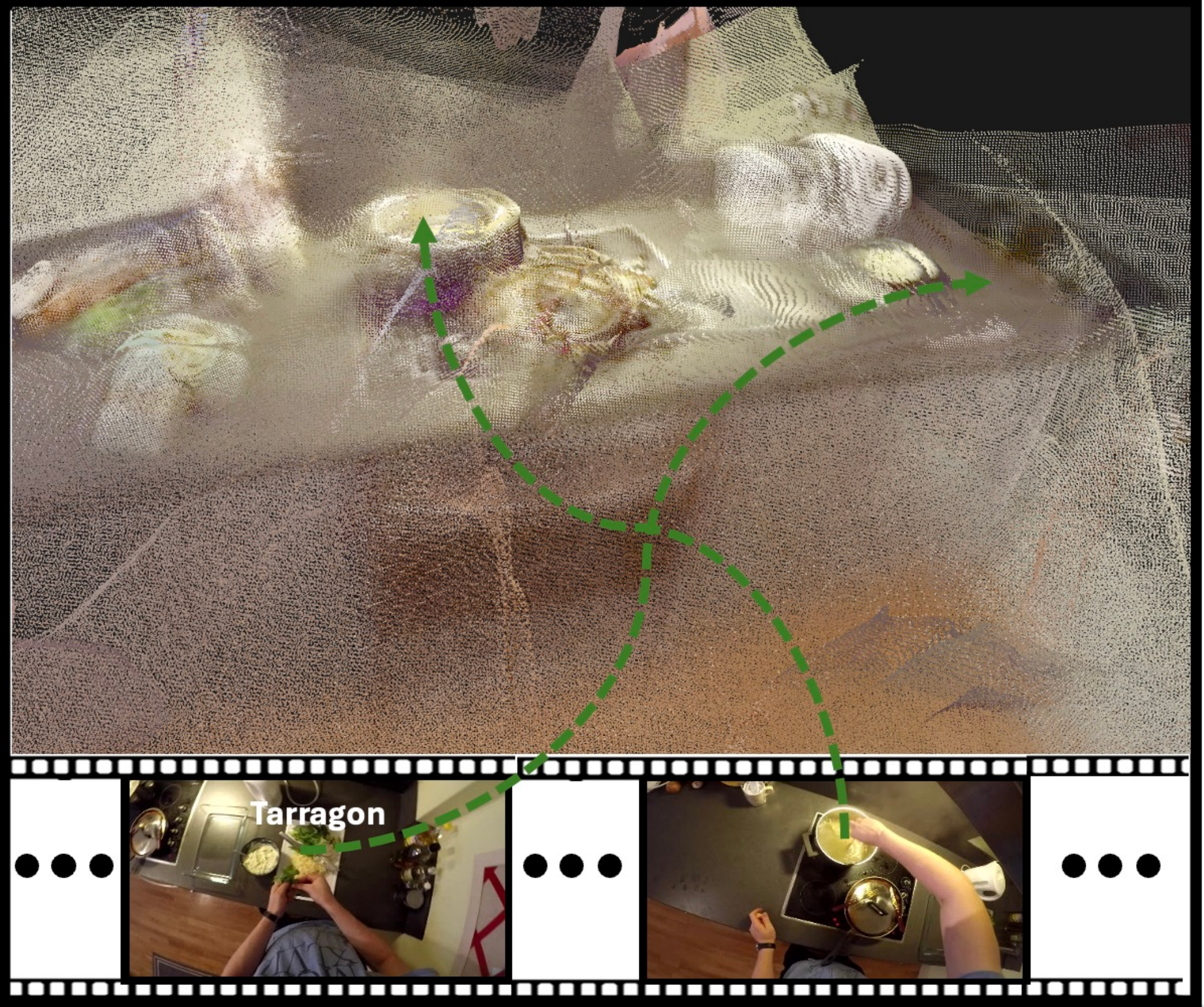}
\vspace{-5pt}
\end{wrapfigure}

\textbf{Question}: Where is the tarragon, and how can the person get to it?

\textbf{Ground-truth}: The tarragon is located on the countertop to the right of the person. The person can reach it by turning to their right and walking towards the countertop.

\textbf{Ours}: The tarragon is on the countertop to the right of the person. The person can reach it by turning to their right and walking towards the countertop. \textcolor{green}{\ding{51}}

\textbf{LLaVA-OV-Qwen2-7B}: The tarragon appears to be on the countertop, specifically on a plate or tray with other chopped vegetables. To get to it, the person would need to move around the kitchen counter and reach towards the area where the tarragon is placed. \textcolor{red}{\ding{55}} 

\textbf{LLaVA-Video-7B-Qwen2}: The tarragon is on the counter in front of the person. The person can get to it by moving closer to the counter and picking up the bowl containing the tarragon. \textcolor{red}{\ding{55}} 

\textbf{Qwen2-VL-7B-Instruct}: The tarragon is located on the kitchen counter. The person can get to it by walking towards the counter and reaching for the tarragon. \textcolor{green}{\ding{51}}
\end{tcolorbox}

\begin{tcolorbox}[
    enhanced,
    breakable,
    colback=yellow!10!white,
    colframe=colorafford,
    title=Furniture Affordance,
    coltitle=black,
    fonttitle=\bfseries,
    listing only,
    listing options={
      basicstyle=\ttfamily\footnotesize,
      breaklines=true,
      showstringspaces=false
    }
]

\begin{wrapfigure}{r}{0.38\linewidth}
\vspace{-5pt}
\includegraphics[width=\linewidth, keepaspectratio]{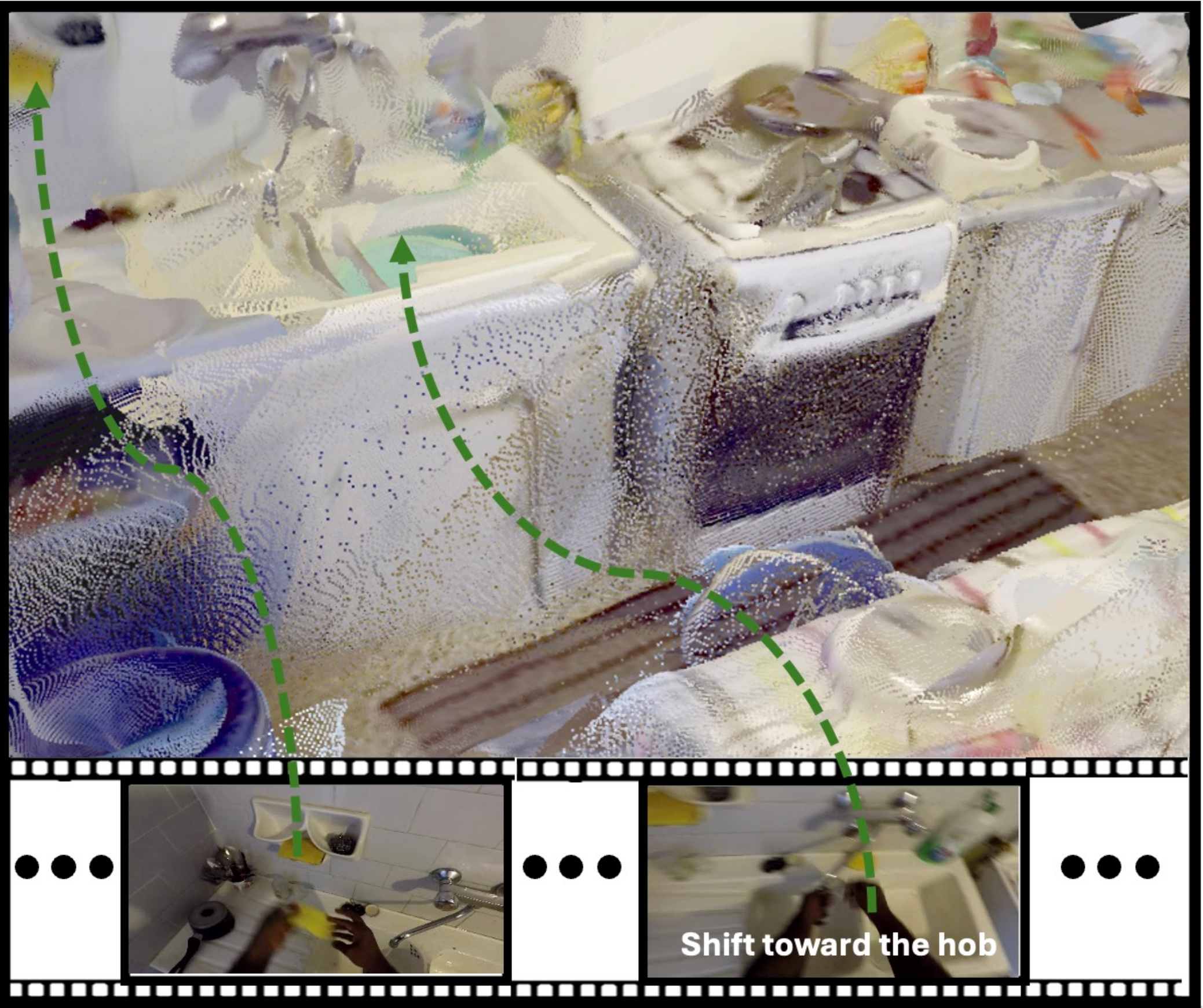}
\vspace{-5pt}
\end{wrapfigure}

\textbf{Question}: Which of the following objects does the person interact with next, given their previous actions and current motion?

A. hob, B. oven, C. fridge

\textbf{Ground-truth}: The person is most likely to interact with the hob next because they have just finished cleaning the sink and are now moving towards the hob, which suggests they might need to use it for cooking or cleaning purposes.

\textbf{Ours}: The person is most likely to interact with the hob next because they have just finished washing dishes and are now moving towards the hob, which is typically used for cooking. \textcolor{green}{\ding{51}}

\textbf{LLaVA-OV-Qwen2-7B}: B. oven. \textcolor{red}{\ding{55}}

\textbf{LLaVA-Video-7B-Qwen2}: B. oven \textcolor{red}{\ding{55}}

\textbf{Qwen2-VL-7B-Instruct}: I don't know. \textcolor{red}{\ding{55}}

\end{tcolorbox}

\newpage
\begin{tcolorbox}[
    enhanced,
    breakable,
    colback=yellow!10!white,
    colframe=colorplan,
    title=Action Planning,
    coltitle=black,
    fonttitle=\bfseries,
    listing only,
    listing options={
      basicstyle=\ttfamily\footnotesize,
      breaklines=true,
      showstringspaces=false
    }
]

\begin{wrapfigure}{r}{0.40\linewidth}
\vspace{-5pt}
\includegraphics[width=\linewidth, keepaspectratio]{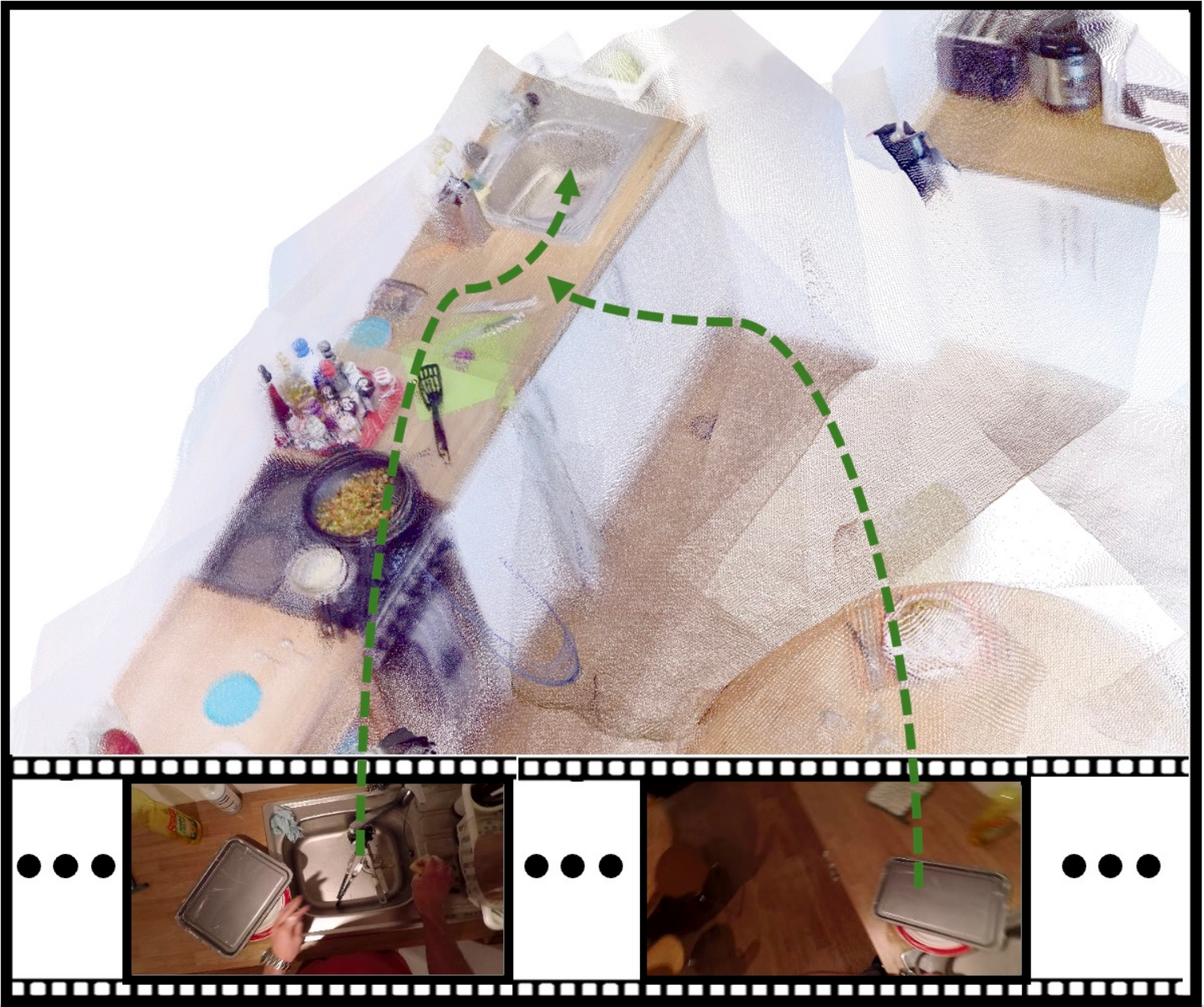}
\vspace{-5pt}
\end{wrapfigure}

\textbf{Question}: We are performing a cooking/assembly task with the following sequence of actions: put down tray, put down sponge, take glass, wash knife.. Based on the video, what should I do next, and how can I get to the place where the next step takes place?

\textbf{Ground-truth}: You have already completed the actions of putting down the tray and the sponge. Now, you are about to take the glass and wash the knife. To do this, move right to the sink and start washing the knife.

\textbf{Ours}: You have already completed the actions of putting down the tray and the sponge, and taking the glass. Now, you need to wash the knife. To do this, move right to the sink and wash the knife there. \textcolor{green}{\ding{51}}

\textbf{LLaVA-OV-Qwen2-7B}: Based on the video, it appears that the person is in the process of cleaning up and preparing for another task. The sequence of actions you've mentioned suggests that the person has already put down a tray and a sponge, taken a glass, and washed a knife. The next step would likely involve using the washed knife to prepare food or assemble something. To determine the next step and where it takes place, let's analyze the last few frames: 1. The person is standing near the sink area. 2. There is a wooden table with a glass and a can nearby. 3. A hand wearing a watch is visible, which might belong to the person performing these tasks. Considering these details, the next logical action could be to use the knife to cut or chop ingredients, which would typically be done on a cutting board. Since the person is already near the sink, they might move towards the wooden table where there is a glass and a can, possibly to find additional ingredients or tools needed for the task. Given the context, the next step would likely involve placing the knife on the table, perhaps on a cutting board if one is available, and then preparing food or assembling an item using the knife. If a cutting board is not present, the person might need to locate one. \textcolor{red}{\ding{55}}

\textbf{LLaVA-Video-7B-Qwen2}: The next step in the cooking/assembly task should be to put the washed knife back into the sink or a designated storage area. To get to the place where this step takes place, the person should move towards the sink area where the knife was washed earlier. The sink is located on the left side of the frame, adjacent to the countertop with the red plate and the pizza. The person can then reach into the sink to place the knife back, ensuring it is properly cleaned and stored. \textcolor{red}{\ding{55}}

\textbf{Qwen2-VL-7B-Instruct}: Based on the sequence of actions provided, the next step after washing the knife is to put down the sponge. The reasoning is that after washing the knife, it is typically appropriate to clean the hands or other utensils used, which would involve using the sponge. To get to the place where the next step takes place, you should head towards the sink area. In the video, the person is seen washing the knife in the sink, and then putting down the sponge. Therefore, you should move towards the sink to continue with the cleaning process. \textcolor{red}{\ding{55}}

\end{tcolorbox}

\subsection{LLM Judge Failure Case}
We use \texttt{ChatGPT-4o} and \texttt{Gemini 2.0 Flash} as automatic judges to evaluate model predictions. However, in some cases, the judges produce incorrect judgments due to rigid comparisons. The example below is marked as incorrect by \texttt{ChatGPT-4o}, even though the prediction and ground truth both correctly identify the fridge as the next object of interaction and provide reasonable justifications.

The only difference lies in the explanation detail: the ground truth states the person placed cheese into the fridge, while the prediction assumes they took cheese and are moving toward the fridge. This type of reasoning difference is speculative and acceptable, as the core intent and predicted outcome remain the same.

\begin{tcolorbox}[
    enhanced,
    breakable,
    colback=yellow!10!white,
    colframe=colorafford,
    title=Furniture Affordance,
    coltitle=black,
    fonttitle=\bfseries,
    listing only,
    listing options={
      basicstyle=\ttfamily\footnotesize,
      breaklines=true,
      showstringspaces=false
    }
]
\textbf{Question}: Which of the following objects does the person interact with next, given their previous actions and current motion?

A. fridge, B. oven

\textbf{Ground-truth}: The person is most likely to interact with the fridge next because they have just placed the mozzarella inside it, and their current movement is towards the fridge, indicating they are likely to retrieve or interact with something else from it.

\textbf{Ours}: The person is most likely to interact with the fridge next because they have just taken a piece of cheese and are now moving towards the fridge, which is a common place to store cheese. \textcolor{red}{\ding{55}}  

\end{tcolorbox}

\subsection{Model Failure Prediction}

\begin{tcolorbox}[
    enhanced,
    breakable,
    colback=yellow!10!white,
    colframe=colorreldir!75,
    title=Relative Direction,
    coltitle=black,
    fonttitle=\bfseries,
    listing only,
    listing options={
      basicstyle=\ttfamily\footnotesize,
      breaklines=true,
      showstringspaces=false
    }
]
\textbf{Question}: Are the hob and freezer on the same side of the person when performing `open drawer'?

\textbf{Ground-truth}: No, the hob is in front of the person, while the freezer is to the left of the person when the person performing `open drawer'.

\textbf{Ours}: Yes, both the hob and freezer are to the left of the person when performing `open drawer'. \textcolor{red}{\ding{55}}  

\end{tcolorbox}

\section{Dataset Details} \label{sec:G}

\subsection{Dataset Statistics}
We provide detailed statistics of our REA dataset.
Table~\ref{tab:rea_stats} reports dataset statistics by tasks, showing the number of question-answer pairs for each task across training and validation splits.
Table~\ref{tab:dataset_video_stats} presents the dataset overview, including video duration, QA counts, average video length, and the number of unique objects and actions.
\vspace{1em}
\begin{table}[h]
\centering
\caption{REA dataset statistics by tasks.}
\vspace{-1em}
\small
\begin{tabular}{lcc}
\hline
\textbf{Task} & \textbf{Train} & \textbf{Validation} \\
\hline
Relative Distance & 4,796 & 300 \\
Relative Direction & 4,765 & 300 \\
Furniture Affordance & 4,192 & 279 \\
Action Planning & 6,500 & 600 \\
Find My Item & 4,118 & 278 \\
\midrule
Total & 24,371 & 1,757 \\
\hline
\end{tabular}
\label{tab:rea_stats}
\end{table}
\vspace{1em}

\begin{table}[h]
\centering
\caption{Dataset Overview.}
\vspace{-1em}
\small
\resizebox{\textwidth}{!}{%
\begin{tabular}{lcccccc}
\toprule
\textbf{Split} & \textbf{Video Duration (hrs)} & \textbf{\# Video IDs*} & \textbf{\# QAs} & \textbf{Avg Video Duration (sec)} & \textbf{\# Unique Objects} & \textbf{\# Unique Actions} \\
\midrule
Train & 221.80 & 152 & 24,371 & 32.76 & 299 & 4,759 \\
Test  & 23.16  & 71 & 1,757  & 47.46 & 94 & 1,309 \\
\bottomrule
\end{tabular}}
\vspace{1mm}
\small{* We refer to the \texttt{video\_id} in EPIC-KITCHENS~\citep{damen2018scaling}.}
\label{tab:dataset_video_stats}
\end{table}

\noindent\textbf{Dataset comparison.} 
Table~\ref{tab:dataset_modality_stats} compares our REA dataset with several related datasets. 
We include Nymeria~\citep{ma2024nymeria} and HD-EPIC~\citep{perrett2025hd}, as well as two additional spatial reasoning benchmarks, VSI-Bench~\citep{yang2024thinking} and All-Angles Bench~\citep{yeh2025seeing}. 
This comparison provides context for the scale and unique properties of our REA dataset, which emphasizes joint spatio-temporal reasoning grounded in both 3D scene structure and egocentric action video.
\vspace{1em}
\begin{table*}[h]
\centering
\caption{Comparison of Spatial Understanding Datasets.}
\vspace{-1em}
\resizebox{\textwidth}{!}{%
\begin{tabular}{lcccccc}
\toprule
\textbf{Dataset} & \textbf{Total Video Duration (hrs)} & \textbf{QA Pairs} & \textbf{RGB Video} & \textbf{3D Modality} & \textbf{Camera Pose} & \textbf{Labelled 3D Environment} \\
\midrule
REA (Ours) & 238.33 & 26.2K & \checkmark & Point Cloud + Multi-view Images & \checkmark & \checkmark \\
HD-EPIC~\citep{perrett2025hd} & 41.3 & 26.5K & \checkmark & 3D Mesh & \checkmark & \checkmark \\
Nymeria~\citep{ma2024nymeria} & 300 & 310.5K & \checkmark & 3D Point Cloud + Sensors & \checkmark & \checkmark \\
VSI-Bench~\citep{yang2024thinking} & N/A & 5K & \checkmark & $\times$ & \checkmark & \checkmark \\
All-Angles Bench~\citep{yeh2025seeing} & N/A & 2.1K & $\times$ & Multi-view Images & \checkmark & $\times$ \\
\bottomrule
\end{tabular}%
}
\label{tab:dataset_modality_stats}
\end{table*}

\subsection{Dataset Generation Pipeline Runtime}

We report the runtime required to generate 5,000 VQA samples for each step in our pipeline, using a single NVIDIA RTX 4090 GPU:

\begin{itemize}
    \item \textbf{Query Video Sampling:} 
    \begin{itemize}
        \item \emph{Relative Direction, Relative Distance:} 1.5 hours
        \item \emph{Action Planning, Furniture Affordance, Find My Item:} 3 hours (requires 7B VLM in the loop)
    \end{itemize}
    \item \textbf{3D Position Estimation:} 10 minutes
    \item \textbf{Spatial Relationship Estimation:} 10 minutes
    \item \textbf{Navigation Movement Extraction:} 20 minutes
    \item \textbf{Scene Reconstruction:} 5 seconds per scene, including saving the output as a \texttt{.glb} file
    \item \textbf{Frame-to-Point Cloud Registration:} Under 4 hours total for 5,000 query videos (batch size = 1) to retrieve corresponding frames from the database.
\end{itemize}

\subsection{QA Templates}
In this section, we present the QA templates we adopt in our data generation pipeline. Note, the curly-braced values (e.g., \{object\_1\}, \{a1\}, \{direction\}) are placeholders. See details in Sec.~\ref{subsec:pipeline}.

\begin{tcolorbox}[
    enhanced,
    breakable,
    colback=yellow!10!white,
    colframe=colorreldir,
    title=Relative Direction,
    coltitle=black,
    fonttitle=\bfseries,
    listing only,
    listing options={
      basicstyle=\ttfamily\footnotesize,
      breaklines=true,
      showstringspaces=false
    }
]

\textbf{Single-object:}

\medskip

\textbf{Q:} Does the hand closer to the \{object\_1\} differ when performing `\{a1\}' and `\{ak\}'?  

\textbf{A:} No, the same hand remains closer to the \{object\_1\} during both `\{a1\}' and `\{ak\}'.  

\textbf{A:} Yes, the hand closer to the \{object\_1\} changes from the \{direction\} of the person to the \{direction\} of the person between `\{a1\}' and `\{ak\}'.

\medskip

\textbf{Q:} Is the \{object\_1\} to the \{direction\} of the person when the person is performing `\{a1\}', and \{format\_direction(direction\_at\_ak)\} when `\{ak\}'? 

\textbf{A:} The \{object\_1\} remains \{format\_direction(direction\_at\_a1)\} during both `\{a1\}' and `\{ak\}'.  

\textbf{A:} Initially, the \{object\_1\} is to the \{direction\} of the person, but as the person moves, it is to the \{direction\} of the person.  

\textbf{A:} At first, the \{object\_1\} appears to the \{direction\} of the person, but after performing `\{ak\}', due to the person's movement, it appears to the \{direction\} of the person.  

\textbf{A:} Relative to the person, the \{object\_1\} changes from being to the \{direction\} of the person to the \{direction\} of the person between `\{a1\}' and `\{ak\}'.

\medskip

\textbf{Multi-object:}

\medskip

\textbf{Q:} Are the \{object\_1\} and \{object\_2\} on the same side of the person when performing `\{a1\}'? 

\textbf{Q:} Is the person facing both the \{anchor\_object\_1\} and \{anchor\_object\_2\} from the same side when performing `\{ak\}'?  

\textbf{A:} Yes, both the \{object\_1\} and \{object\_2\} are to the \{direction\} of the person during `\{a1\}'.

\textbf{A:} No, the \{object\_1\} is to the \{direction\} of the person, while the \{object\_2\} is to the \{direction\} of the person during `\{a1\}'.

\end{tcolorbox}

\begin{tcolorbox}[
    enhanced,
    breakable,
    colback=yellow!10!white,
    colframe=colorreldist,
    title=Relative Distance,
    coltitle=black,
    fonttitle=\bfseries,
    listing only,
    listing options={
      basicstyle=\ttfamily\footnotesize,
      breaklines=true,
      showstringspaces=false
    }
]

\textbf{Single-object:}

\medskip

\textbf{Q:} Does the person move closer to the \{object\_1\} between `\{a1\}' and `\{ak\}'?  

\textbf{Q:} Does the person move away from the \{object\_1\} between `\{a1\}' and `\{ak\}'?  

\textbf{Q:} Does the person end up closer to the \{object\_1\} after performing `\{ak\}'?  

\textbf{Q:} Is the person closer to the \{object\_1\} when `\{a1\}' or when `\{ak\}'?  

\textbf{Q:} During which action is the person closest to the \{object\_1\}?

\textbf{A:} The person moves closer to the \{object\_1\} from `\{a1\}' to `\{ak\}'.  

\textbf{A:} The person starts off farther from the \{object\_1\} at `\{a1\}', but ends up closer to it after `\{ak\}'.  

\textbf{A:} The person approaches the \{object\_1\} while moving from `\{a1\}' to `\{ak\}'.  

\textbf{A:} The person moves further away from the \{object\_1\} from `\{a1\}' to `\{ak\}'.  

\textbf{A:} The person starts off closer to the \{object\_1\} at `\{a1\}', but ends up farther from it after `\{ak\}'.  

\textbf{A:} The person moves away from the \{object\_1\} while moving from `\{a1\}' to `\{ak\}'.

\medskip

\textbf{Multi-object:}

\medskip

\textbf{Q:} During `\{a1\}', is the person closer to the \{object\_1\} than to the \{object\_2\}?  

\textbf{Q:} During `\{a1\}', would it be easier for the person to access the \{object\_1\} or the \{object\_2\}?

\textbf{A:} Yes, the person is closer to the \{object\_1\} than to the \{object\_2\} when performing `\{a1\}'.  

\textbf{A:} No, the person is closer to the \{object\_2\} than to the \{object\_1\} when performing `\{a1\}'.  

\textbf{A:} The person is at a similar distance from both the \{object\_1\} and the \{object\_2\} when performing `\{a1\}'.  

\textbf{A:} The person's relative distance to \{object\_1\} and \{object\_2\} is unclear when performing `\{a1\}'.

\end{tcolorbox}

\begin{tcolorbox}[
    enhanced,
    breakable,
    colback=yellow!10!white,
    colframe=colorfindmy,
    title=Find My Item,
    coltitle=black,
    fonttitle=\bfseries,
    listing only,
    listing options={
      basicstyle=\ttfamily\footnotesize,
      breaklines=true,
      showstringspaces=false
    }
]

\textbf{Q:} Where is the \{object\_1\}, and how can the person get to it?  

\textbf{Q:} After performing \{action\_name\}, where did the person leave the \{object\_1\}, and how can it be reached?  

\textbf{Q:} Would it be closer for the person to bring the \{object\_1\} to the \{anchor\_object\_1\} or to the \{anchor\_object\_2\}?

\medskip

\textbf{A:} Answers for this task are free-form generations produced by a VideoLLM. Responses may describe the object's location, surrounding context, and suggested navigation steps, depending on the scene and queried action. See Sec.~\ref{subsec:videollm_prompts} for more details.

\end{tcolorbox}

\begin{tcolorbox}[
    enhanced,
    breakable,
    colback=yellow!10!white,
    colframe=colorafford,
    title=Furniture Affordance Prediction,
    coltitle=black,
    fonttitle=\bfseries,
    listing only,
    listing options={
      basicstyle=\ttfamily\footnotesize,
      breaklines=true,
      showstringspaces=false
    }
]

\textbf{Q:} Considering the person’s previous actions and current movement, which object will they most likely interact with next?  

\textbf{Q:} Which of the following objects does the person interact with next, given their previous actions and current motion?  

\textbf{Q:} Based on what the person has done so far and how they are moving now, which nearby object is the person preparing to interact with?

\medskip

\textbf{A:} Answers are free-form generations from a VideoLLM, typically referring to a plausible next object interaction such as “fridge,” “microwave,” “sink,” or “hob.” These responses are selected based on the temporal progression and motion cues present in the scene. See Sec.~\ref{subsec:videollm_prompts} for more details.

\end{tcolorbox}

\begin{tcolorbox}[
    enhanced,
    breakable,
    colback=yellow!10!white,
    colframe=colorplan,
    title=Action Planning,
    coltitle=black,
    fonttitle=\bfseries,
    listing only,
    listing options={
      basicstyle=\ttfamily\footnotesize,
      breaklines=true,
      showstringspaces=false
    }
]

\textbf{Q:} We are performing a cooking or assembly task with the following sequence of actions: <action\_1>, <action\_2>, ..., <action\_i>. Based on the video, what should I do next, and how can I get to the place where the next step takes place?

\medskip

\textbf{A:} Answers are free-form generations from a VideoLLM, which may describe the predicted next action (e.g., “close salt,” “put down spatula”) and the spatial guidance for reaching the appropriate location (e.g., “move to the stove,” “turn toward the counter on the left”). These answers require reasoning over the temporal context and understanding of task progression. See Sec.~\ref{subsec:videollm_prompts} for more details.

\end{tcolorbox}

\subsection{Answer Generation Prompts}
\label{subsec:videollm_prompts}
In this section, we present the instructions we use to prompt the VideoLLM to construct free-form ground-truth answers during QA generation.

\begin{tcolorbox}[
    enhanced,
    breakable,
    colback=yellow!10!white,
    colframe=colorfindmy,
    title=Find My Item,
    coltitle=black,
    fonttitle=\bfseries,
    listing only,
    listing options={
      basicstyle=\ttfamily\footnotesize,
      breaklines=true,
      showstringspaces=false
    }
]

\textbf{If question\_type is ``location'':}

\medskip

\textbf{<images>} You are given a short video showing the action: \{action\_name\}.  
In the video, the person places the object: \{object\_1\}.

\begin{itemize}
  \item The video only shows the \textbf{past action} — the moment the object was last placed.
  \item You, the assistant, are \textbf{not in the video}, and the person's current position is unknown.
  \item You are told the object is now located \textbf{\{direction\_phrase\}} from the person’s current position. This direction is accurate and must be used in the response.
\end{itemize}

\textbf{Question:} Where is the \{object\_1\}, and how can the person get to it?

\textbf{Your answer must:}
\begin{itemize}
  \item Describe the surroundings around the object at the last moment it was visible (based only on the video)
  \item Use the known direction ``\{direction\_phrase\}'' to state where the object is now and how the person can reach it
  \item Not guess or infer directions from the video
  \item Not mention the video directly
  \item Not invent room layouts or paths
  \item Be one fluent, natural English sentence
\end{itemize}

\medskip
\textbf{If question\_type is ``after action'':}

\medskip

\textbf{<images>} You are given a short video showing the action: \{action\_name\}.  
In this video, the person places the object: \{object\_1\}.

\begin{itemize}
  \item The video shows only the \textbf{past action} of placing the object.
  \item You are \textbf{not currently in the video}.
  \item The person’s current position is \textbf{after the video ends} and not visible.
  \item You are told the object is now located at: \textbf{\{direction\_phrase\}}, and this direction must be used exactly as given.
\end{itemize}

\textbf{Question:} After performing \{action\_name\}, where did the person leave the \{object\_1\} and how to reach it?

\textbf{Your answer must:}
\begin{itemize}
  \item Describe the surroundings where the object was placed at the end of the action (based only on the video)
  \item Use the known direction ``\{direction\_phrase\}'' to describe where the object is now and how the person can reach it
  \item Not infer direction from the video
  \item Not use generic phrases like ``to the right'' unless they match ``\{direction\_phrase\}''
  \item Not mention the video directly or invent room layouts
  \item Be one fluent and natural English sentence
\end{itemize}

\end{tcolorbox}

\begin{tcolorbox}[
    enhanced,
    breakable,
    colback=yellow!10!white,
    colframe=colorafford,
    title=Furniture Affordance,
    coltitle=black,
    fonttitle=\bfseries,
    listing only,
    listing options={
      basicstyle=\ttfamily\footnotesize,
      breaklines=true,
      showstringspaces=false
    }
]

\textbf{<images>} You are given information about a person and their surroundings:

\begin{itemize}
  \item \textbf{Previous actions performed by the person}: \{previous\_actions\_text\}
  \item \textbf{Movement relative to nearby objects}: \{movement\_text\}
  \item \textbf{Available object options}: \{options\_text\}
\end{itemize}

Your task is to generate a fluent, natural English sentence that answers the following question:

\textbf{``Which object will the person most likely interact with next?''}

\begin{itemize}
  \item The answer should indicate that the person is most likely to interact with the \{groundtruth\_anchor\_object\}.
  \item Do not mention the list of options directly in your answer.
  \item Explain naturally why the person is approaching or likely to interact with the \{groundtruth\_anchor\_object\}, based on their actions and movement.
  \item Keep the response concise and human-like.
  \item Do not repeat the question or include unrelated commentary.
\end{itemize}

\end{tcolorbox}

\begin{tcolorbox}[
    enhanced,
    breakable,
    colback=yellow!10!white,
    colframe=colorplan,
    title=Action Planning,
    coltitle=black,
    fonttitle=\bfseries,
    listing only,
    listing options={
      basicstyle=\ttfamily\footnotesize,
      breaklines=true,
      showstringspaces=false
    }
]

\textbf{<images>} You are an assistant that generates a detailed, natural-language answer in the second person, describing progress and the next step in an egocentric video. The data you have is:

\begin{itemize}
  \item \textbf{Video ID}: \{video\_id\}
  \item \textbf{Video Frames Range}: from \{start\_frame\} to \{end\_frame\}
  \item \textbf{Two key frames for reference}: \{", ".join(img\_input\_list)\}
  \item \textbf{Actions in the overall sequence}: \{", ".join(all\_actions)\}
  \item \textbf{Actions completed in the video so far}: \{", ".join(completed\_actions)\}
  \item \textbf{Next action to perform}: \{next\_action\}
  \item \textbf{Motion data for how to perform the next action}: \{movement\_type\}
\end{itemize}

Your task:
\begin{enumerate}
  \item Acknowledge which actions have already been completed, based on the video.
  \item Infer the user's immediate next step from the provided \texttt{next\_action}.
  \item Describe how the user will physically carry out this next action, considering:
  \begin{itemize}
    \item Movement type (e.g., forward, backward, left, right, or stand still if movement is minimal)
    \item \{extra\_rotation\_instruction\}
  \end{itemize}
  \item Respond in a single, natural-sounding sentence or short paragraph, addressing the user as ``you'' (second-person perspective), like an on-the-spot assistant.
  \item If the location for the next action is not evident from the video, reference the point cloud to determine where the user needs to go. Then provide navigation instructions (e.g., “move right” or “turn left toward the counter”) so the user can reach the correct spot and perform the action.
\end{enumerate}

Please generate a concise, coherent answer that incorporates these details, focusing on telling the user what they have already done and how to perform the next action.

\end{tcolorbox}

\subsection{Licenses}
EPIC-KITCHENS~\citep{damen2018scaling}, EPIC-FIELDS~\citep{EPICFields2023}, VISOR~\citep{VISOR2022} are licensed under  \href{https://creativecommons.org/licenses/by-nc/4.0/}{CC BY-NC 4.0}. We thank the authors of these datasets for providing high-quality annotations, which form the basis of our work.

\section{Extended Limitations}
\label{sec:limit}

For tasks such as \task{colorreldir}{\emph{Relative Direction}} and \task{colorreldist}{\emph{Relative Distance}}, we adopt fixed templates to generate candidate answers. While this enables efficient evaluation, it introduces a risk of overfitting to the specific answer formats rather than encouraging diverse and natural responses. To mitigate this, future work could incorporate LLM in the loop to paraphrase or refine the answer templates, promoting more robust generation and improving generalization to non-templated question-answer pairs.

\section{LLM Usage} \label{sec:llm_usage}
While preparing this work, we used a large language model (LLM) to assist with language editing. The core research, experimental design, and all scientific claims remain our original work.
Beyond editing, LLMs were also employed in the data generation pipeline to generate and refine question–answer pairs, 
and further served as automated judges for the evaluation of results.

\end{document}